\documentclass[10pt,twocolumn,letterpaper]{article}

\usepackage[pagenumbers]{cvpr}
\usepackage{cuted}
\usepackage{currfile}
\usepackage{caption}

\usepackage{tabularx,array}
\usepackage{booktabs}
\usepackage{enumitem}
\usepackage{algorithm}
\usepackage{algorithmic}
\usepackage{amsmath,amssymb}
\usepackage{graphicx}
\graphicspath{{fig/}{./}}
\newcommand{\SuppRecPos}{12}
\newcommand{\SuppRecTot}{12}
\newcommand{\SuppGatePos}{12}
\newcommand{\SuppGateTot}{12}
\newcommand{\SuppMetaPos}{12}
\newcommand{\SuppMetaTot}{15}
\newcommand{\SuppLLMCalls}{50}
\newcommand{\SuppLLMUSD}{2.53}
\newcommand{\SuppLLMMaxUSD}{0.71}
\newcommand{\SuppPilotGPUh}{213.4}
\newcommand{\SuppPilotRuns}{21}
\newcommand{\Suppnumfix}{$+1.50$ {\tiny $[-0.33, +3.38]$}}
\newcommand{\Suppllmfix}{$+2.65$ {\tiny $[+0.73, +4.58]$}}

\newcommand{\SuppExtRLeighteenThree}{$+7.23$ {\tiny $[+5.55, +8.90]$}}
\newcommand{\SuppExtRLthirtysixThree}{$+1.87$ {\tiny $[+0.40, +3.33]$}}

\newcommand{\mref}[1]{\cref{#1}}
\definecolor{cvprblue}{rgb}{0.21,0.49,0.74}
\usepackage[pagebackref,breaklinks,colorlinks,allcolors=cvprblue]{hyperref}
\usepackage[capitalize,nameinlink]{cleveref}

\title{Gated Recursive Self-Improvement for Robot Policies: From Self-Practice to Program Adaptation}

\author{
Zeyu Wang$^{1}$ \quad Xian Wang$^{2}$ \quad Yuxuan Li$^{3}$ \quad Pengbin Feng$^{4}$ \quad Yuhang Yao$^{5}$ \quad Shangding Gu$^{6}$ \quad Hongshan Lin$^{7}$\\
Dangen She$^{8}$ \quad Junjie Wu$^{8}$ \quad Donghan Li$^{9}$ \quad Yuyi Zhi$^{2}$ \quad He Dong$^{7}$ \quad Hao Li$^{10}$ \quad Hongjian Zhang$^{11}$\\
Zongyi Li$^{7}$ \quad Hanxin Chen$^{10}$ \quad Sitan Liu$^{7}$ \quad Guogeng Li$^{10}$ \quad Yujia Zhang$^{12}$ \quad Tianyu Shi$^{13}$\\[0.6em]
{\small $^{1}$University of California, Los Angeles \quad $^{2}$Zhejiang University \quad $^{3}$Georgia Institute of Technology \quad $^{4}$Amazon}\\
{\small $^{5}$Carnegie Mellon University \quad $^{6}$University of California, Berkeley \quad $^{7}$Independent Researcher \quad $^{8}$Li Auto Inc.}\\
{\small $^{9}$University of Pittsburgh \quad $^{10}$University of California, San Diego \quad $^{11}$The Chinese University of Hong Kong, Shenzhen}\\
{\small $^{12}$University of Alberta \quad $^{13}$McGill University}
}

\begin{document}
\maketitle
\begin{abstract}
We study how a robot policy can keep improving over many rounds of practice with simulator success feedback, without costly new demonstrations.
Existing self-training methods can improve policies from such feedback, but repeated practice alone does not guarantee continued progress.
Each retained policy generates the next round's experience, while the training program determines how that experience is used; poor updates and ineffective programs can therefore hinder subsequent learning.
We present a gated recursive self-improvement (RSI) framework for robot policies, in which a language model adapts the training program from the records of past rounds and a gate retains a candidate policy with its program only if it matches or beats its parent on the same validation placements, redrawn each round.
We apply RSI to a 7B OpenVLA-OFT policy on LIBERO-Goal and LIBERO-Spatial, measuring success on 1{,}000 newly sampled placements per suite that training and selection never use.
With a working GRPO program kept fixed, RSI raises success over twelve rounds and ends 7.3 to 8.6 points above the same loop with data from the frozen initial policy.
With a behaviour-cloning program that degrades the policy when every update is accepted, the gate keeps success 8.4 to 15.5 points above the ungated loop after six rounds, and program adaptation raises success further.
Starting from the same program, RSI also exceeds a tuned GRPO baseline on LIBERO-Goal by 2.2 points after six rounds, using about half the GPU-hours spent tuning and training that baseline.
These results suggest judging self-training both by the policy it retains and by whether its training program can still produce useful updates.
\end{abstract}

\begin{figure}[t]
  \centering
  \includegraphics[width=\linewidth]{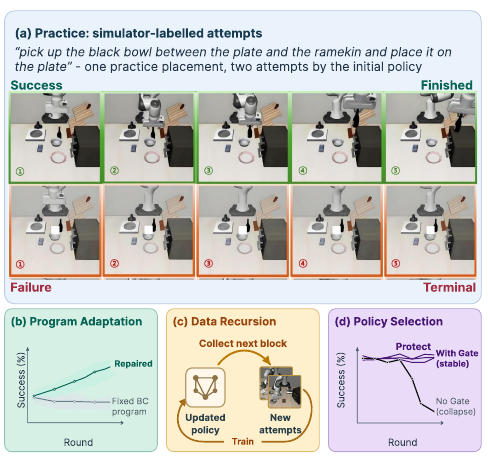}
    \caption{
    \textbf{Recursive self-improvement for a robot policy.}
    (a) \textbf{Practice:} Initial policy rollouts on LIBERO-Spatial, labelled as success or failure by the simulator.
    (b) \textbf{Program adaptation:} Meta-level repair improves an ineffective BC policy on LIBERO-Goal.
    (c) \textbf{Data recursion:} Updating the rollout policy improves success rates over using the frozen initial policy.
    (d) \textbf{Policy selection:} The gate prevents harmful updates, while accepting all updates causes performance to collapse from 58.5\% to 7\%.
    }
  \label{fig:teaser}
\end{figure}
\section{Introduction}
\label{sec:intro}
Collecting demonstrations is costly in robot learning. A vision-language-action (VLA) policy can learn from its own interactions without further demonstrations, using binary success feedback from simulation~\cite{li2025simplevlarl,liu2025rl4vla}. The practical question is how to turn repeated practice into a sequence of useful policy updates.

A retained policy both performs the task and generates data for further learning. Updating it therefore changes task performance and the experience available in the next round. Selecting policy updates and revising the program that trains them also shape this process. Studying these decisions separately can explain how self-training sustains improvement and restores learning from an ineffective starting program.

\begin{figure*}[t]
  \centering
  \includegraphics[width=0.95\textwidth]{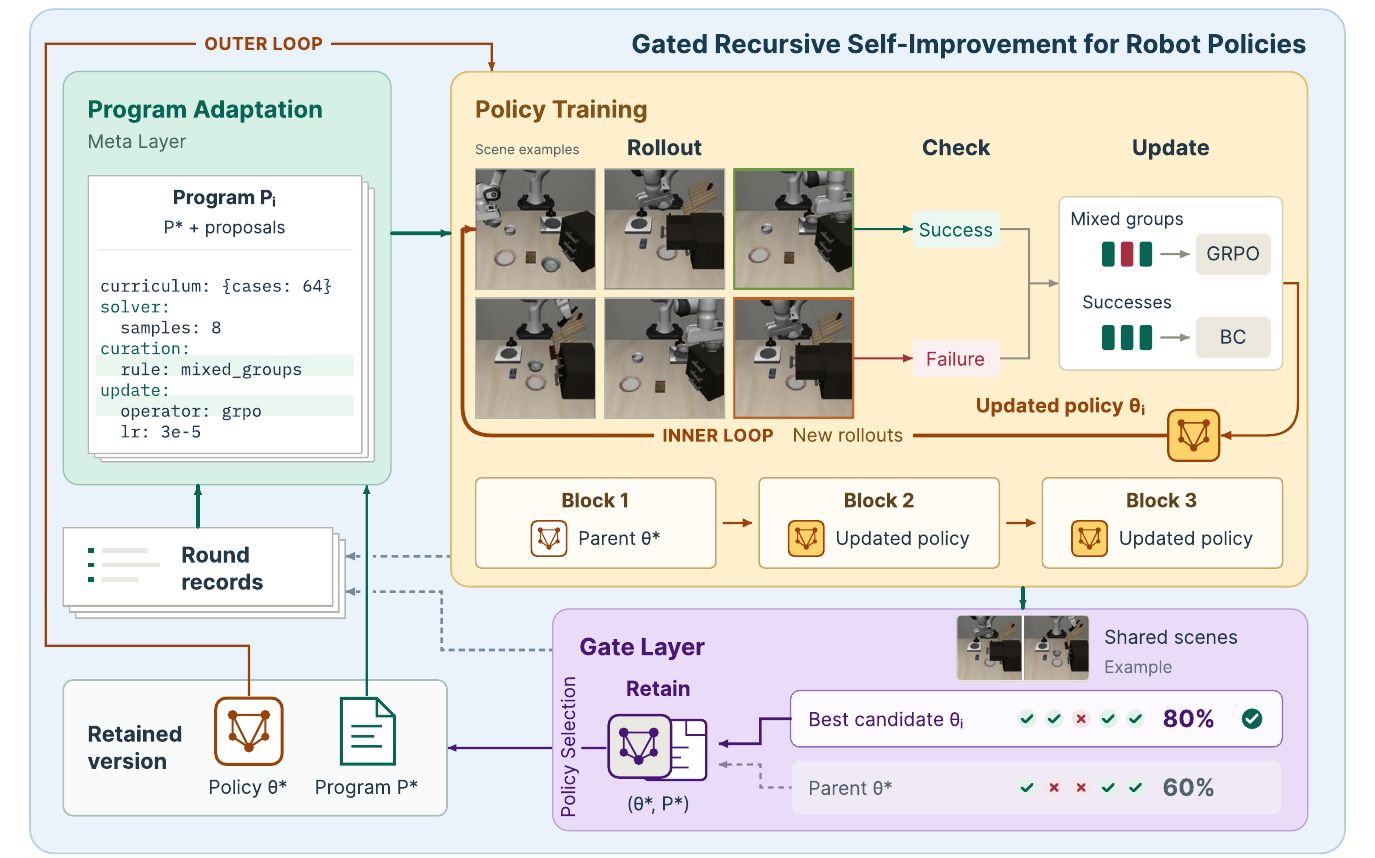}
  \caption{\textbf{Gated recursive self-improvement for robot policies.} 
  The inner loop performs three training blocks, updating the policy from verifier-labelled rollouts using GRPO or BC. The updated policy generates each subsequent block’s data. A gate compares candidates with their parent on paired held-out placements, retaining the best candidate and its program if it matches or exceeds the parent’s success rate; otherwise, the parent pair is kept. The meta layer uses round records to propose alternatives alongside the retained program.}
  \label{fig:framework}
\end{figure*}

Our recursive self-improvement (RSI) framework places data generation, policy selection and program adaptation in a common implementation (\cref{fig:framework}). Controlled comparisons freeze the data-generating policy, shuffle success labels, accept every update, or keep the training program fixed. Numeric search and a language model provide alternative ways to adapt the program.

Experiments on LIBERO-Goal and LIBERO-Spatial distinguish two starting conditions (\cref{fig:teaser,tab:components}). With a working GRPO program, updating the data generator supports continued learning on unseen scenes. With a behaviour-cloning program that degrades under repeated updates, the gate preserves initial performance and program adaptation enables further gains. We also compare the adapted loop with tuned one-shot RL, counting the resources used to select RL's training configuration.

Fresh scenes provide a common measure of improvement across these studies. LIBERO~\cite{liu2023libero} uses fixed initial placements that can also be used for training. Our protocol excludes test states from the learning loop and evaluates on both the official states and 1{,}000 fresh scenes. A separate analysis of a released VLA-RL checkpoint estimates a $+16.8$-point contribution associated with training on the test states, about a fifth of its reported score. We present the component comparisons in \cref{sec:rq1,sec:rq2,sec:rq3}, followed by compute cost and the effect of evaluation states in \cref{sec:rl,sec:rq4}.

\section{Related work}
\label{sec:related}

\paragraph{Learning from self-generated experience.} STaR~\cite{zelikman2022star}, ReST~\cite{gulcehre2023rest} and ReST-EM~\cite{singh2024restem} fine-tune language models on their own verified rationales. SPIN~\cite{chen2024spin} and self-rewarding models~\cite{yuan2024selfrewarding} also use model-generated outputs for further training; Absolute Zero~\cite{zhao2025absolutezero} generates tasks and verifies solutions through code execution. Group-relative policy optimisation is a common update rule for RL with verifiable rewards~\cite{shao2024deepseekmath,guo2025deepseekr1}. Our component comparisons examine how the source of this experience and its success labels contribute to repeated policy improvement. The on-policy analysis of~\cite{kim2026coverage} predicts early saturation with samples from a frozen initial policy, as observed in our twelve-round non-recursive control.

\paragraph{Adapting the training procedure.} LLM-driven search has been applied to programs and training configurations with large search spaces and inexpensive evaluation~\cite{novikov2025alphaevolve,lu2024aiscientist,hu2024adas}. Meta-Harness~\cite{lee2026metaharness} exposes earlier candidates' code, scores and traces; EvoTrainer~\cite{chen2026evotrainer} adds diagnostics and a memory of failed branches; LLMZero~\cite{fang2026llmzero} analyses signals at training checkpoints. Our workspace-based proposer also retains records for later proposals. We compare numeric and language-model proposals from working and failing programs, where each candidate requires policy training and rollout-based evaluation. More broadly, updating configurations from noisy, expensive simulator rollouts without an explicit analytical objective parallels work in likelihood-free and amortized inference, which learns parameter mappings and posterior uncertainty directly from simulated trajectories~\cite{zhang2025likelihood, zhang2025dimension, zhu2026aim}.

\paragraph{Robot learning and program improvement.} OpenVLA~\cite{kim2024openvla} and OpenVLA-OFT~\cite{kim2025oft} provide our policy family. SimpleVLA-RL~\cite{li2025simplevlarl}, RL4VLA~\cite{liu2025rl4vla} and related methods~\cite{lu2025vlarl} improve VLA policies through on-policy RL and binary success rewards. We use the SimpleVLA-RL stack and its released checkpoints. Concurrent approaches construct robot programs from execution feedback, including a code library requiring no model at run time~\cite{hu2026coap}, a validated skill hierarchy~\cite{wen2026roborsi}, and programs compiled from exploration~\cite{huang2026rax}. Our controlled comparisons separate improvements from adapting the training program and from updating policy weights.

\paragraph{Preregistered evaluation.} Preregistration specifies hypotheses and analyses before their results are observed~\cite{nosek2018preregistration,bertinetto2021preregistration}. We use it to specify comparisons, effect thresholds and analysis procedures.

\section{Problem Formulation}
\label{sec:problem}

We study robot policy improvement from a demonstration-trained policy $\theta_0$ and an initial training program $P_0$, without additional demonstrations. As the policy learns, its behavior changes the experience available for subsequent updates; records of completed training attempts can inform changes to the training program.

\paragraph{Task performance.}
A task instance $c=(\ell,x_0)$ comprises a language instruction $\ell$ and an initial environment configuration $x_0$. In our VLA instantiation, $\pi_\theta$ maps the instruction, camera images and proprioceptive observations to action chunks. Interaction with environment $E$ produces a trajectory $\tau$, labelled by a fixed external success rule $v(c,\tau)\in\{0,1\}$. For a fixed target distribution $\mu$, performance is
\begin{equation}
J(\theta)=\mathbb{E}_{c\sim\mu}\,
\mathbb{E}_{\tau\sim p_E^{\mathrm{eval}}(\cdot\mid c,\theta)}
[v(c,\tau)],
\end{equation}
where $p_E^{\mathrm{eval}}$ includes policy execution and environment dynamics under a fixed evaluation protocol. Our goal is to improve $J$ relative to $\theta_0$, measured on instances excluded from training and policy selection.

\paragraph{Improvement process.}
A program $P\in\mathcal{P}$ specifies instance sampling, rollout collection, training-data selection and policy updates, with $\mathcal{P}$ the declared space of admissible programs. The pair $(\theta,P)$ defines a version from which learning continues. Completed training and validation records support decisions about which version to retain and whether to revise its program. Records from rejected candidates remain available for subsequent program proposals. These decisions rely on finite observations of performance. Program changes may alter practice-instance sampling, while the target distribution $\mu$ and success rule $v$ remain fixed.

\section{Framework}
\label{sec:method}
Each round trains candidate policies and selects the version used to continue learning (\cref{fig:framework,alg:loop}). All candidates start from the same retained weights. Each is trained under the current program or a proposed alternative, then compared with the parent on the same validation instances. The selected policy generates the next round's experience, and candidate records inform later program proposals.

\subsection{Learning from self-generated experience}
\label{sec:experience}
Within a \emph{training block}, policy $\theta$ attempts sampled instances $K$ times each. Program $P$ specifies how these attempts are collected and selected for a LoRA update. The selection rule depends on the update operator. For GRPO~\cite{shao2024deepseekmath}, \texttt{mixed\_groups} retains all attempts from instances whose success rate falls strictly within a specified band. Groups with uniform outcomes have zero group-centred advantage. For behaviour cloning (BC), \texttt{verified} retains only successful attempts.

For the bowl placement in \cref{fig:teaser}(a), the initial policy produces both a successful and an unsuccessful attempt. BC trains on the successful trajectory; GRPO uses the outcomes within a retained group to assign relative advantages. The gate evaluates the updated policy on other placements.

A \emph{round} trains each candidate for three successive blocks. Within a candidate, each block continues from the previous block's adapter and starts with a fresh optimiser. In the recursive configuration, the current candidate collects each block's experience, beginning with $\theta^\star$ in the first block. This is the \emph{data recursion}: the generator changes as the policy learns. The \texttt{nonrec} control instead uses the frozen initial policy $\theta_0$ to collect every block's experience, while continuing to update the learner from $\theta^\star$.

\begin{algorithm}[t]
\caption{Gated recursive self-improvement}
\label{alg:loop}
\begin{algorithmic}[1]
\REQUIRE $\theta_0$, $P_0$, success rule $v$, rounds $R$, candidates $c$
\STATE $(\theta^\star, P^\star) \leftarrow (\theta_0, P_0)$
\FOR{$r = 1 \ldots R$}
  \STATE $\{P_i\}_{i \le c} \leftarrow \{P^\star\} \cup \textsc{Propose}(\text{records}_{<r})$ \hfill $\triangleright$ meta
  \STATE draw fresh $G_r$; $g^\star \leftarrow$ success of $\theta^\star$ on $G_r$
  \FOR{each $P_i$}
     \STATE $\theta_i \leftarrow \theta^\star$
     \FOR{block $= 1 \ldots 3$}
        \STATE $\theta_{\text{gen}} \leftarrow \theta_i$ (recursive) or $\theta_0$ (\texttt{nonrec})
        \STATE attempts $\leftarrow$ $K$ rollouts of $\theta_{\text{gen}}$ on \textsc{Sample}$(P_i)$
        \STATE labels $\leftarrow v$ (\texttt{shuffled}: permuted per instance)
        \STATE $\theta_i \leftarrow \textsc{Update}(\theta_i, \textsc{Select}(\text{attempts}, P_i), P_i)$
     \ENDFOR
     \STATE $g_i \leftarrow$ success of $\theta_i$ on $G_r$ (greedy)
  \ENDFOR
  \STATE $i^\star \leftarrow \arg\max_i g_i$; keep $(\theta_{i^\star}, P_{i^\star})$ if $g_{i^\star} \ge g^\star$
\ENDFOR
\STATE \textbf{return} $(\theta^\star, P^\star)$ \hfill $\triangleright$ tests read after all runs end
\end{algorithmic}
\end{algorithm}

\subsection{Selecting policy updates}
\label{sec:selection}
The \emph{gate} compares each trained candidate with its parent on the same newly sampled validation instances. We use 20 instances per task, for 200 episodes, with greedy decoding. Re-seeding the environment for each episode gives the parent and candidates identical initial configurations. The highest-scoring candidate and its program are retained if its success rate is at least the parent's; otherwise the parent version is retained.

Each round resamples the validation set and re-evaluates the parent, giving a paired comparison on fresh validation instances. We measure the variability of this comparison in \cref{sec:rq2}. The \texttt{nogate} control accepts every candidate. Audit and sealed-test sets provide independent evaluations of the retained policies (\cref{sec:protocol}).

\subsection{Adapting training programs}
\label{sec:meta}
At the start of each round, the \emph{meta level} reads records from completed rounds and proposes $c{-}1$ programs in $\mathcal{P}$, without seeing the current validation scores. We train one candidate under each proposed program and one under the current program $P^\star$, all from the same parent weights. The selected policy retains the program that produced it. In the \texttt{fixed} arm, the program stays unchanged while the policy continues to learn.

We compare three proposers. Numeric search (\texttt{full}) randomly mutates declared parameters, favouring those whose previous changes improved the gate score. The plain language-model proposer (\texttt{llm}, Claude Sonnet) receives the current program, neutral edit descriptions and earlier candidates' records. These contain program changes, gate scores, acceptance decisions and per-block statistics: sampled success, the share of instances with mixed outcomes and data counts. The model writes a diagnosis from these records before proposing a program.\footnote{The prompt also specified the first and last training losses; a field-name mismatch left these fields out of the records received by this proposer.}

The workspace-based proposer (\texttt{llm\_harness}) lets the same model inspect a fuller record over several turns, following~\cite{lee2026metaharness,chen2026evotrainer}. The files contain candidate histories, per-task scores and a ledger linking edits to gate-score changes. They also include a frozen memory of earlier runs, a playbook prepared before launch and the proposer's notes. The model can list, read and search these files; its prompt states the measured gate noise.

The \emph{knob} space contains ten hyperparameters; the \emph{harness} space adds nine procedural choices, including the policy-gradient ratio, loss aggregation and BC likelihood temperature (supplementary material). The edit space was assembled from manual changes informed by round records. Descriptions explain what each option changes and leave the choice of setting to the proposer.

\paragraph{Proposer inputs and compute budget.} The proposer operates on program configurations and round records. Audit and test results, policy weights and episodes remain outside its inputs. Programs pass the numeric-mutation validator; invalid proposals fall back to numeric search and are counted. An optional \emph{compute cap} limits each candidate to the initial program's rollout and optimiser budget.

\section{Experimental setup}
\label{sec:setup}

\begin{table}[t]
\centering
\caption{\textbf{Component ablation results.}
Success-rate differences (pp) with 95\% CIs.}
\label{tab:components}
\scriptsize
\setlength{\tabcolsep}{4pt}
\renewcommand{\arraystretch}{1.25}

\begin{tabular}{@{}lcc@{}}
\toprule
\textbf{Comparison (setup)}
& \textbf{Official}
& \textbf{Fresh} \\
\midrule

\multicolumn{3}{@{}l}{\textbf{LIBERO-Goal}} \\
\addlinespace[2pt]

Recursion (GRPO/12)
& \shortstack{$+7.84$ \\ {\tiny $[5.22,10.49]$}}
& \shortstack{$+7.30$ \\ {\tiny $[5.47,9.20]$}} \\

Gate strictness (GRPO/12)
& \shortstack{$+0.89$ \\ {\tiny $[-1.11,2.89]$}}
& \shortstack{$+1.67$ \\ {\tiny $[0.00,3.30]$}} \\

Gate (BC/6)
& \shortstack{$+13.60$ \\ {\tiny $[10.53,16.76]$}}
& \shortstack{$+15.47$ \\ {\tiny $[13.20,17.77]$}} \\

LLM meta-level (BC/6)
& \shortstack{$+2.65\smash{^\dagger}$ \\ {\tiny $[0.73,4.58]$}}
& \shortstack{$+4.33$ \\ {\tiny $[2.57,6.10]$}} \\

\midrule
\multicolumn{3}{@{}l}{\textbf{LIBERO-Spatial}} \\
\addlinespace[2pt]

Recursion (GRPO/12)
& \shortstack{$+8.67$ \\ {\tiny $[6.76,10.69]$}}
& \shortstack{$+8.63$ \\ {\tiny $[7.17,10.10]$}} \\

Gate strictness (GRPO/12)
& \shortstack{$+0.29$ \\ {\tiny $[-1.51,2.13]$}}
& \shortstack{$+1.17$ \\ {\tiny $[-0.07,2.40]$}} \\

Gate strictness (GRPO/6)
& \shortstack{$+0.73\smash{^\dagger}$ \\ {\tiny $[-0.77,2.25]$}}
& \shortstack{$+0.35\smash{^\dagger}$ \\ {\tiny $[-0.62,1.32]$}} \\

Gate (BC/6)
& \shortstack{$+13.00$ \\ {\tiny $[10.44,15.60]$}}
& \shortstack{$+8.37$ \\ {\tiny $[6.67,10.07]$}} \\

LLM meta-level (BC/6)
& \shortstack{$+0.53\smash{^\dagger}$ \\ {\tiny $[-1.05,2.13]$}}
& \shortstack{$+3.60\smash{^\dagger}$ \\ {\tiny $[2.50,4.72]$}} \\

\bottomrule
\end{tabular}

\vspace{3pt}
\parbox{\columnwidth}{\footnotesize
Comparisons: Recursion = fixed $-$ nonrec;
Gate strictness = guard $-$ fixed;
Gate = fixed $-$ nogate;
LLM meta-level = full RSI $-$ fixed.
Setup: initial recipe/rounds.
$^\dagger$ Seeds 0--3 as registered; other entries seeds 1--3.
}
\end{table}

\subsection{Benchmarks, policies and starting points}
\label{sec:starting}
We evaluate our proposed RSI framework on LIBERO-Goal and LIBERO-Spatial~\cite{liu2023libero}. Goal contains ten goal-conditioned tasks in one scene; Spatial contains ten pick-and-place tasks with different layouts of the same objects. The separate checkpoint analysis uses LIBERO-10's ten long-horizon tasks. Our policy is OpenVLA-OFT~\cite{kim2025oft} (7B, action chunks of 8), initialised from supervised fine-tuning on one demonstration per task~\cite{li2025simplevlarl}. Initial success is about 59\% on LIBERO-Goal and 60\% on LIBERO-Spatial.

Each action chunk contains 56 discrete tokens (256 bins), produced in one parallel-decoding forward pass. Rollouts sample at temperature $T$, also used for GRPO log-probabilities; gates and tests use greedy decoding. Our implementation uses the OpenVLA-OFT model code from SimpleVLA-RL\cite{li2025simplevlarl}, with custom rollout and GRPO/BC update workers and LoRA rank $r{=}32$.

The main component study uses a fixed GRPO training recipe informed by pilot experiments and chosen before the main runs. The recipe uses a learning rate of $3\times10^{-5}$, temperature-matched policy ratios, and episode-balanced loss. Each block collects eight attempts on each of 64 instances at temperature 1.6, then performs 100 optimiser steps with batch size 64. Each round comprises three blocks, with fresh paired gate evaluations determining whether to accept the updated policy.

Comparisons use seeds 1--3 on both suites, except entries marked $^\dagger$ in \cref{tab:components}. The recursion and gate-strictness experiments start from the same fixed GRPO recipe and run for twelve rounds, with program adaptation disabled. They compare recursive against non-recursive data generation, and strict gating against a relaxed guard-rail gate, respectively.

The gate-removal and self-repair experiments start from the same failing BC recipe and run for six rounds. The gate-removal comparison keeps recursion enabled and the training recipe fixed, changing only whether updates can be rejected by the gate. The self-repair comparison keeps recursion and strict gating enabled and compares the fixed BC recipe against full RSI with LLM-based program adaptation. The LLM can modify training parameters and switch the update algorithm, including from BC to GRPO. The fixed-recipe arms train one candidate per round, while the LLM arm trains two. \Cref{tab:components} reports the component ablation results on LIBERO-Goal and LIBERO-Spatial under both official and fresh evaluation protocols.

\subsection{Experimental design and controls}
\label{sec:controls}
\begin{table}[t]
\centering
\footnotesize
\caption{\textbf{Experimental controls.}
Fixed is the baseline within each starting recipe.}
\label{tab:controls}
\setlength{\tabcolsep}{5pt}
\renewcommand{\arraystretch}{1.15}
\begin{tabular}{@{}lcccc@{}}
\toprule
Arm & Recursion & Gate & Meta & Candidates \\
\midrule
\texttt{fixed}  & On  & Strict  & Off & 1 \\
\texttt{nonrec} & Off & Strict  & Off & 1 \\
\texttt{nogate} & On  & None    & Off & 1 \\
\texttt{guard}  & On  & Relaxed & Off & 1 \\
\texttt{llm}    & On  & Strict  & LLM & 2 \\
\bottomrule
\end{tabular}
\end{table}

\Cref{tab:controls} summarises the component controls. In \texttt{fixed}, the training recipe stays fixed while the policy learns. In \texttt{nonrec}, the initial policy generates training data while the learner continues to update.
The strict gate requires the candidate to match or exceed its parent's success rate on paired gate evaluations. The \texttt{guard} arm permits a decrease of at most 5 percentage points relative to the parent, while also
requiring the candidate to remain no more than 5 points below the initial policy. The \texttt{nogate} arm removes rejection based on performance degradation.
Starting recipes and training schedules are described in \cref{sec:starting}.
For the system-level comparison on LIBERO-Goal, we compare full LLM RSI with two tuned GRPO baselines, RL-18 and RL-36. Starting from the same initial policy, these baselines perform 18 or 36 sequential on-policy training blocks
without intermediate gates. We report the policy retained by a final gate. Unlike RSI, which starts from the failing BC recipe, both baselines use a tuned GRPO recipe. We report execution GPU-hours separately from total costs
including the one-time RL recipe-search cost of 213.4 GPU-hours. LLM inference costs are excluded from the reported GPU-hours.
Detailed results and analyses of these comparisons
are presented in \cref{sec:results}.

\subsection{Evaluation protocol}
\label{sec:protocol}
\paragraph{Evaluation sets.} We evaluate on two held-out test sets. \textbf{Official} uses the benchmark's 50 fixed initial states per task, giving 500 episodes per evaluation. \textbf{Fresh} uses 1,000 newly sampled initial configurations from the same task suite, excluded from training, gating, and auditing. Official measures performance on the standard benchmark states, while Fresh tests whether performance extends to new configurations of the same tasks.
Gate and audit evaluations use separate initial configurations that do not overlap the test sets. An additional 200-episode audit measures performance outside the gate set without influencing policy selection. Neither Official nor Fresh results are used to select policies or update training recipes.

\paragraph{Final policies and evaluation repeats.} We report the final retained policy from each run. For the component comparisons, Official results average three evaluations per seed, while Fresh results use the
first 1,000-episode evaluation per seed. For the system-level comparison with tuned RL, we average all available final evaluations within each seed.
RL-18 seed 1 retains the initial policy and has one Official evaluation.
We then average success rates equally across seeds 1--3 (seeds 1 and 2 for the continuation in \cref{sec:rl}).

\paragraph{Statistical analysis.} We report 95\% confidence intervals (CIs) for paired success-rate differences using a seed-averaged, per-episode paired bootstrap with 10,000 draws. Pairing preserves the correspondence between the two methods' outcomes on the same evaluation instances. These intervals quantify evaluation-instance uncertainty conditional on the observed training seeds. They do not fully capture variability across independent training runs.
An interval entirely above zero indicates a positive difference under this procedure, whereas an interval containing zero does not establish a clear direction. All differences are reported in percentage points.
\section{Results}
\label{sec:results}

\begin{figure*}[t]
  \centering
  \includegraphics[width=\textwidth,
  trim={0 10mm 0 0},
  clip]{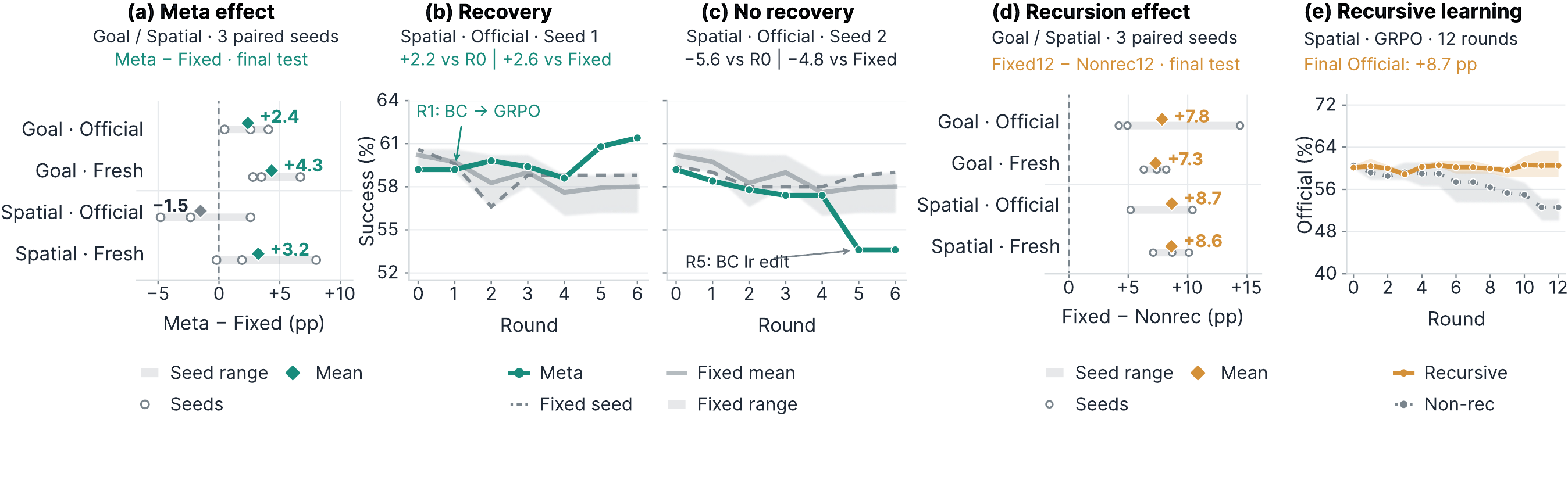}
  \caption{\textbf{Program adaptation and data recursion.} (a) Meta level minus \texttt{fixed} on the final tests, seeds 1--3. (b, c) Official success by round on Spatial, seeds 1 and 2, against \texttt{fixed} (\cref{tab:selfrepair}). (d) \texttt{fixed} minus \texttt{nonrec} after twelve rounds. (e) Spatial, GRPO: recursive versus non-recursive data generation.}
  \label{fig:selfrepair}
\end{figure*}

We evaluate recursive self-improvement on LIBERO-Spatial
and LIBERO-Goal through three controlled component studies: \textbf{Program Adaptation}, \textbf{Data Recursion}, and \textbf{Policy Selection}.
Beyond these components, we address two broader research questions:
\begin{itemize}
    \item \textbf{Compute Efficiency:}
    How does RSI compare with conventional RL in terms
    of performance and computational cost?
    \item \textbf{Evaluation Protocol:}
    How sensitive are measured improvements to the evaluation protocol?
\end{itemize}

\subsection{Program Adaptation}
\label{sec:rq1}
We examine whether meta-level program adaptation can repair an ineffective
BC training program on LIBERO-Goal and LIBERO-Spatial.
All runs use data recursion and a fresh paired gate over six rounds,
starting from a BC recipe with a sampling temperature of 1.6
and unit-temperature likelihoods.
We compare a fixed program (one candidate per round) with
meta-level adaptation (two candidates per round) on the four
registered seeds (0--3). Both suites use LLM-based adaptation, with an additional
numeric-search fallback for Spatial.

\indent\textbf{Finding 1: Meta-level repair raises Fresh success on both suites.}
Adaptation improves Goal Official by $+2.65$ pp $[+0.73, +4.58]$ and Spatial Fresh by $+3.60$ $[+2.50, +4.72]$, above the registered $+3$ threshold; Spatial Official changes by $+0.53$ $[-1.05, +2.13]$ (Goal Fresh: \cref{tab:components}).
These effects include additional candidate search,
as adaptation evaluates twice as many candidates per round.

To understand the variation on Spatial, we examine two contrasting
cases in \cref{fig:selfrepair}(b--c) and \cref{tab:selfrepair}.
In Seed 1, the meta-level switches from BC to GRPO in Round 1,
improving final Official success by $+2.6$ pp over fixed.
In Seed 2, the meta-level retains BC and modifies task sampling
and learning rate. Despite a higher gate score for the Round 5
candidate (68.0\% vs.\ 62.5\%), final Official success falls
$4.8$ pp below fixed.
In seed 1 the change accepted at the gate carries over to the independent test; in seed 2 it does not.

\begin{table}[t]
    \centering
    \small
    \caption{\textbf{Spatial program repair cases.}
    Selected accepted program changes across two seeds.
    Final Official differences average three evaluations
    per retained policy.}
    \label{tab:selfrepair}

    \setlength{\tabcolsep}{3pt}
    \renewcommand{\arraystretch}{1.10}

    \begin{tabular*}{\linewidth}{
        @{\extracolsep{\fill}}lp{0.59\linewidth}r@{}
    }
        \hline
        Round & Key accepted changes
        & \mbox{Official $\Delta$} \\
        \hline

        \multicolumn{2}{@{}l}{
            \textbf{Seed 1}\quad (Recovery)
        } & $+2.6$~pp \\[2pt]

        R1
        & \raggedright
          BC $\to$ GRPO; mixed-group curation;
          token-level loss
        & \\[5pt]

        \multicolumn{2}{@{}l}{
            \textbf{Seed 2}\quad (No recovery)
        } & $-4.8$~pp \\[2pt]

        R3 & task reweighting & \\

        R5 & LR $10^{-5} \to 3\times10^{-5}$ & \\[2pt]

        \hline
    \end{tabular*}
\end{table}

\subsection{Data Recursion}
\label{sec:rq2}
We examine whether data recursion improves self-training on LIBERO-Goal and LIBERO-Spatial. All runs use a fixed GRPO training program and a fresh paired gate over twelve rounds. We compare recursive data generation using the latest policy with non-recursive generation using a frozen initial policy across three paired seeds. Both configurations update the learner, differing only in the policy used to generate training experience.

\indent\textbf{Finding 2: Data recursion consistently
outperforms frozen data generation, with gains that
depend on the evaluation distribution.}
As shown in \cref{fig:selfrepair}(d) and
\cref{tab:recursion}, recursion outperforms the
non-recursive control by $+7.30$--$+8.67$ pp across
both suites and evaluations, with all paired 95\% CIs
above zero. Relative to R0, recursion largely preserves
Official performance ($+0.40$--$+0.93$ pp), while
improving Fresh success by $+3.54$ pp on Goal and
$+7.38$ pp on Spatial. In contrast, the non-recursive
control degrades across all four settings
(\cref{tab:recursion}).

\begin{table}[t]
\centering
\small
\setlength{\tabcolsep}{3pt}
\caption{\textbf{Data recursion over 12 rounds.}
Changes relative to R0 and final paired advantages
(pp; 95\% CI). R0 changes use single Official
evaluations or the shared Fresh reference;
final Official advantages use three evaluations
per seed.}
\label{tab:recursion}
\resizebox{\columnwidth}{!}{%
\begin{tabular}{@{}lrrr@{}}
\toprule
Suite / Eval. &
Rec. $-$ R0 &
Non-rec. $-$ R0 &
Rec. $-$ Non-rec. \\
\midrule
Goal / Official & +0.93 & -5.80 &
\shortstack{$\mathbf{+7.84}$\\
{\scriptsize $[5.22,10.49]$}} \\
Goal / Fresh & +3.54 & -3.76 &
\shortstack{$\mathbf{+7.30}$\\
{\scriptsize $[5.47,9.20]$}} \\
Spatial / Official & +0.40 & -7.93 &
\shortstack{$\mathbf{+8.67}$\\
{\scriptsize $[6.76,10.69]$}} \\
Spatial / Fresh & +7.38 & -1.25 &
\shortstack{$\mathbf{+8.63}$\\
{\scriptsize $[7.17,10.10]$}} \\
\bottomrule
\end{tabular}}
\end{table}

\begin{figure}[t]
  \centering
  \includegraphics[width=\columnwidth]{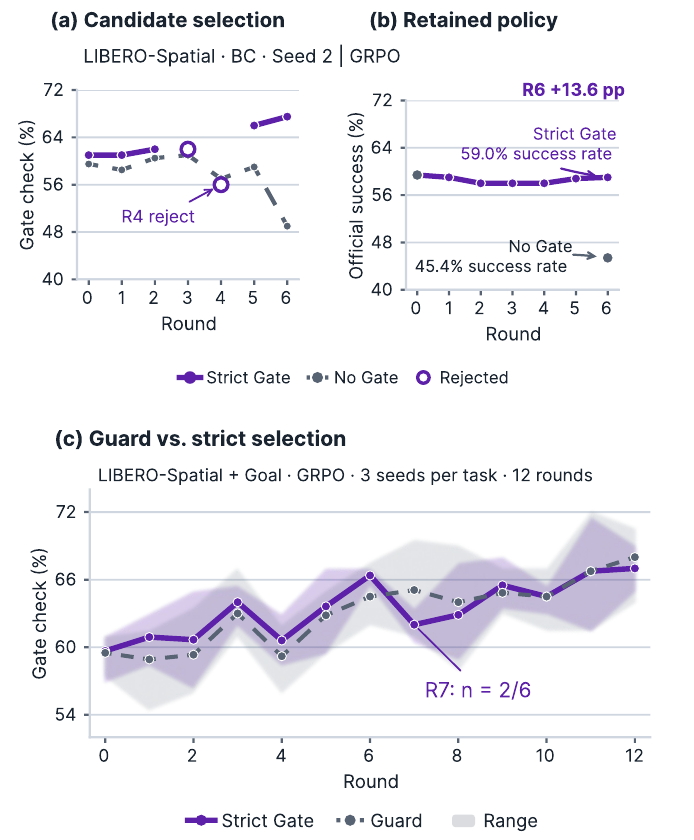}
  \caption{\textbf{Policy selection.} (a, b) Spatial, BC, seed 2: gate-check success of each candidate (open circles: rejected) and Official success of the retained policy, with and without the gate. (c) Strict gate versus relaxed guard, twelve GRPO rounds, three seeds per suite.}
  \label{fig:gates}
\end{figure}
\subsection{Policy Selection}
\label{sec:rq3}
We examine whether an acceptance gate prevents harmful policy updates during self-training. Starting from an ineffective BC training program,
we compare a strict fresh paired gate with always accepting updates over six rounds on LIBERO-Goal and LIBERO-Spatial, using three paired seeds and equal training compute.

\indent\textbf{Finding 3: The acceptance gate prevents
performance collapse by rejecting harmful updates.} \cref{fig:gates}(a,b) illustrates this effect on Spatial (Seed 2). At R3 and R4, the gated
loop encounters candidates with gate-check scores similar to those in ungated training, but rejects both updates and retains the previous policy. By R6, the gated policy maintains $59.0\%$ Official success, while ungated training drops to $45.4\%$.

As shown in \cref{tab:gate}, strict gating limits the mean Official decline on Spatial to $2.20$ pp, compared with $13.80$ pp without the gate. This suggests that the gate primarily protects existing performance rather than repairing an ineffective training program.

More severe failures occur in additional ungated BC runs. On Goal, one run drops to $5.5\%$ mean Gate/Audit success, while a Spatial run reaches only $1.2\%$ Official success after six rounds. In the matched Spatial run, strict gating retains $55.13\%$ Official success, illustrating its protective role.

We also examine gate strictness over 12 rounds of GRPO training
(\cref{fig:gates}(c)). A relaxed guard allowing drops of up to 5 pp produces similar gate-check trends to strict selection, with only small differences in final Official performance.

\begin{table}[t]
\centering
\small
\caption{\textbf{Effect of gate selection.}
Mean Official success across three seeds on
LIBERO-Spatial (BC, six rounds).
Changes are in percentage points.}
\label{tab:gate}
\begin{tabular}{@{}lrrr@{}}
\toprule
Selection & R0 (\%) & R6 (\%) & $\Delta$ (pp) \\
\midrule
Strict Gate & 60.20 & 58.00 & -2.20 \\
No Gate     & 59.33 & 45.53 & -13.80 \\
\bottomrule
\end{tabular}
\end{table}

\subsection{Compute Efficiency}
\label{sec:rl}
\begin{figure}[t]
  \centering
  \includegraphics[width=\linewidth]{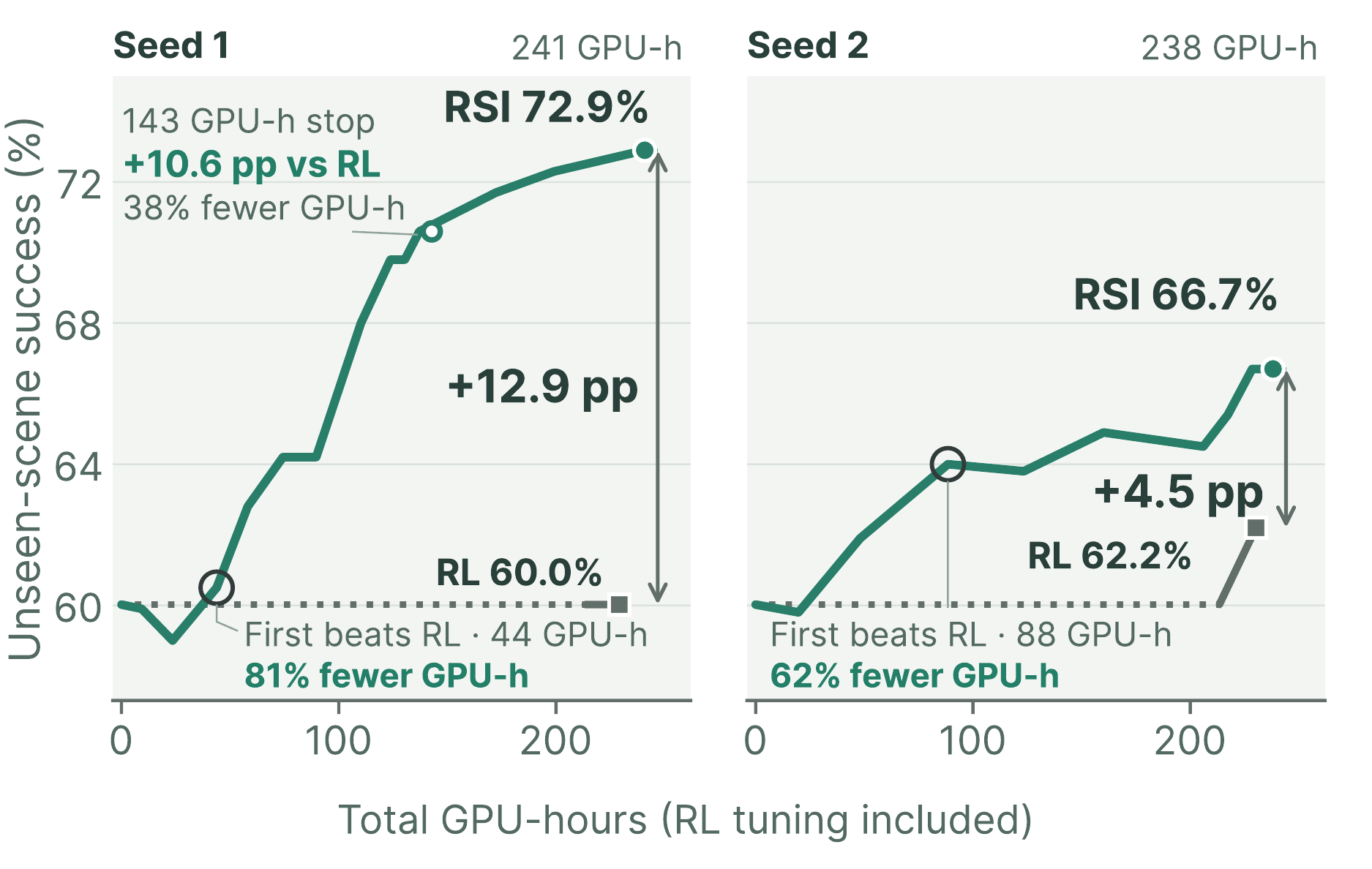}
  \caption{\textbf{Performance and compute cost, including RL tuning} (LIBERO-Goal, 1{,}000 unseen scenes). RL spends 213 GPU-hours choosing its configuration (dotted), then trains 18 blocks; for seed 1 the gate did not keep its update. Green: the loop up to about RL's total GPU-hours; rings: first version above RL; small dot: registered stop. Averaged over the two seeds, at the registered stop the loop ends $+7.6$ $[+5.6, +9.6]$ points above tuned RL at 17\% fewer GPU-hours (\cref{sec:rl}).}
  \label{fig:cost}
\end{figure}

We compare RSI with conventional RL on both performance and total GPU-hours, including the cost of selecting RL's training configuration. On LIBERO-Goal, RL tuning requires 21 pilot runs ($213.4$ GPU-hours), bringing the total cost of RL-18 to $229.9$ GPU-hours. RSI starts from the ineffective BC program without pilots. \Cref{fig:cost} follows the retained policy of each run against total GPU-hours.

\indent\textbf{Finding 4: RSI overtakes tuned RL early and ends above it at lower total cost.} The loop first exceeds RL-18 on fresh scenes after 44 GPU-hours on seed 1 and 88 on seed 2 (rings in \cref{fig:cost}), less than two fifths of RL's total. After six rounds ($118.1$ GPU-hours, seeds 1--3), it exceeds RL-18 by $+2.20$ pp
$[+0.86,+3.53]$ on Fresh evaluations, with comparable Official success ($59.78\%$ vs.\ $59.11\%$).

A preregistered continuation keeps the adapted program fixed. We report it for seeds 1 and 2. The third registered seed is excluded: its practice success fell from 58\% to 32\% and then 10\% in the first continuation rounds while its validation score stayed at 65\%, unlike seeds 1 and 2 (supplementary material). At the registered stop, using 190 GPU-hours on average against RL's 230 (17\% fewer), the loop finishes $+7.6$ points $[+5.6, +9.6]$ above RL-18 on fresh scenes (official test $+4.0$ $[+1.4, +6.6]$). Against RL trained twice as long (RL-36, 36 blocks, $239.6$ GPU-hours including tuning) the difference is $+2.5$ $[+0.8, +4.3]$.

\subsection{Evaluation Protocol}
\label{sec:rq4}
\begin{figure}[t]
  \centering
  \includegraphics[width=\linewidth]{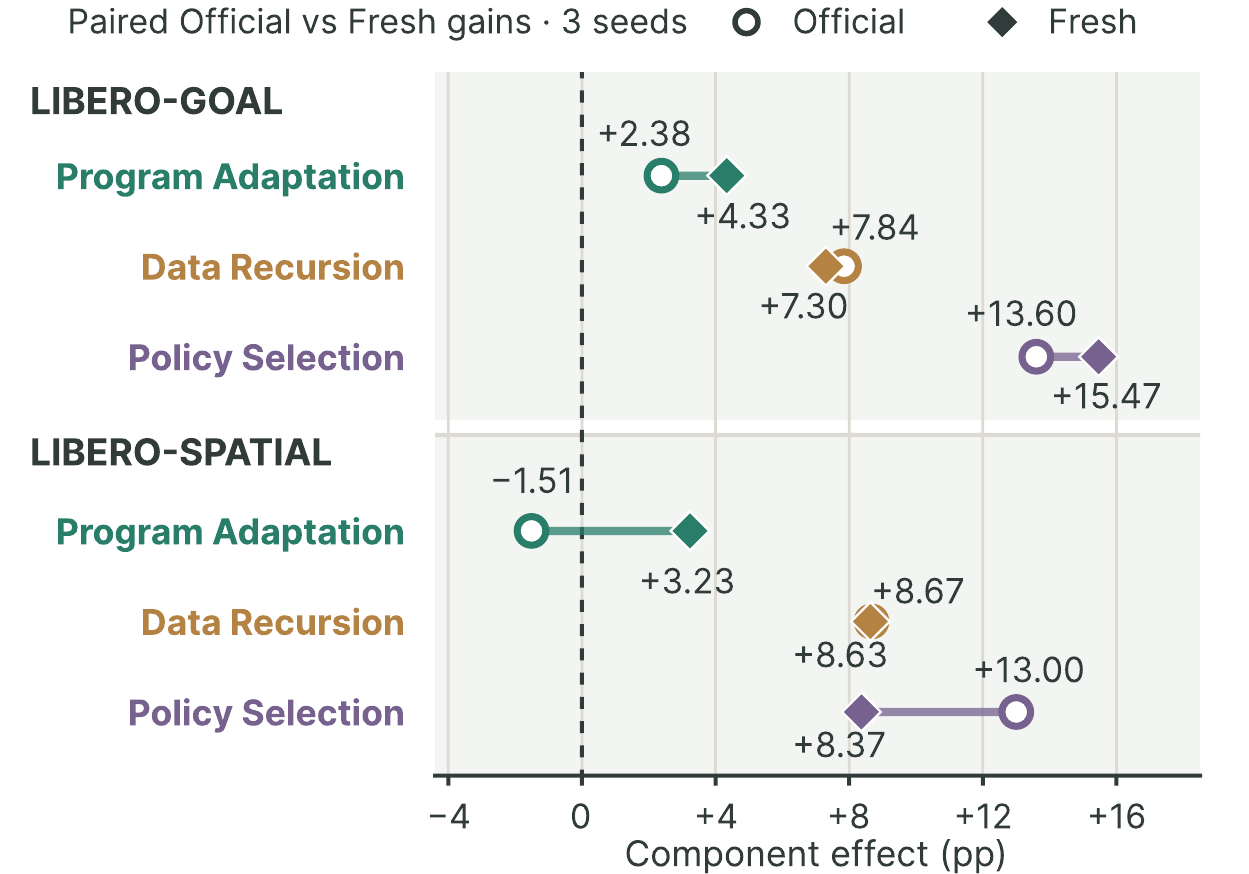}
  \caption{\textbf{Official versus Fresh evaluation.} Component effects (pp) on the official test states (open circles) and on 1{,}000 fresh placements (diamonds), seeds 1--3 (\cref{tab:components}).}
  \label{fig:inflation}
\end{figure}

We examine how evaluation protocols affect measured improvements on LIBERO-Goal and LIBERO-Spatial, comparing Official test states with independently sampled Fresh placements.

\indent\textbf{Finding 5: Measured improvements depend on the evaluation protocol.} As shown in \cref{fig:inflation}, Program Adaptation on Spatial changes from $-1.51$ pp on Official to $+3.23$ pp on Fresh (seeds 1--3). Data Recursion shows similar gains across both evaluations, while Policy Selection varies in magnitude. Reusing test states for training widens the difference between the two protocols. A released SimpleVLA-RL checkpoint trained on LIBERO-10's 50 fixed test states per task scores 87.4\% on the official states and 76.0\% on 500 fresh placements, against 14.6\% and 20.0\% for its SFT start. Subtracting the SFT gap from the RL gap gives $+16.8$ points $[+9.8, +23.4]$, about a fifth of the reported score.

\section{Discussion}
\label{sec:discussion}

Our findings highlight the complementary roles of
the three RSI components. Program adaptation can
help recover underperforming policies by repairing
ineffective training procedures. Data recursion
improves learning by generating new experience
with the updated policy rather than a frozen
data generator. Policy selection protects against
harmful updates but cannot repair an ineffective
training program on its own. Together, these
components allow task feedback to guide both
policy updates and future training decisions.

Program adaptation also changes how compute is
spent by combining policy learning with training
program search. On LIBERO-Goal, RSI achieves
higher Fresh success at lower total GPU-hours
when RL's pilot tuning cost is included.
The saving comes from not needing pilot runs.

\section{Conclusion}
\label{sec:conclusion}

Our experiments clarify the roles of three components in recursive self-improvement for robot policies. Program adaptation can repair ineffective training procedures, data recursion improves learning through updated experience, and policy selection protects against harmful updates. On LIBERO-Goal, RSI achieves higher Fresh success than tuned RL at lower total GPU-hours when RL's tuning cost is included. Finally, differences between Official and Fresh evaluations highlight the importance of evaluating self-improvement on unseen states.

\section*{LLM Usage Statement}

Large language models are part of the method studied here: the \texttt{llm} meta level (\cref{sec:meta}) calls Claude Sonnet 5.5 to propose edits to the training program; its inputs and outputs are logged and released with the run records.

In the research and writing, we used Claude Code, an AI coding assistant (Claude Opus 5.5 and Claude Fable 5.1/5.2), to write and debug experiment and analysis code, launch and monitor runs, and compute the registered reads. We used Claude Code and ChatGPT to draft, revise, and polish text across the paper.
Figures were created using AI-assisted Streamlit scripts and subsequently refined in Figma, including layout, styling, and annotations.

The authors designed the study, fixed each registered claim and its analysis in a preregistration before its data existed, reviewed the code, numbers, figures, and text, and take full responsibility for the paper.

{
    \small
    \bibliographystyle{ieeenat_fullname}
    \bibliography{refs}
}
\clearpage
\appendix
\setcounter{table}{0}\setcounter{figure}{0}
\renewcommand{\thetable}{S\arabic{table}}\renewcommand{\thefigure}{S\arabic{figure}}
\renewcommand{\theHtable}{S\arabic{table}}\renewcommand{\theHfigure}{S\arabic{figure}}

\section*{Supplementary Material}
\addcontentsline{toc}{section}{Supplementary Material}

\noindent This supplement gives the implementation details, the meta level's edit space, prompts and decisions, the per-seed values behind every comparison in the paper, the registration of each comparison, the cost accounting of \cref{sec:rl}, and example rollouts. References without an ``S'' (\eg, \cref{tab:components}, \cref{sec:rl}) point to the main paper. A rollout video of the starting policy and of a self-improved policy on the same five scenes (\cref{sec:rollouts}) is on the project page, \url{https://gated-rsi.pages.dev}, which will also link the code release: the framework, the registered analysis scripts, the preregistration with all dated addenda, and the run records behind every number (\cref{sec:code}).

\section*{Reading guide}
\label{sec:guide}
Short answers to questions a reader may have, with pointers to the evidence.

\paragraph{Why does success start near 60\%?} The policy is fine-tuned on one demonstration per task (\cref{sec:starting}); LIBERO provides 50 per task. No test state is used for training, gating or program adaptation. \cref{sec:rq4} measures how much training on the 50 official test states adds to a released checkpoint.

\paragraph{Are the effects consistent across seeds?} The paper's CIs condition on the training seeds; \cref{tab:perseed} gives every seed. Over both suites and both tests, recursion is positive in \SuppRecPos{} of \SuppRecTot{} seed-level differences, the gate in \SuppGatePos{} of \SuppGateTot{}, and the LLM meta level in \SuppMetaPos{} of \SuppMetaTot{}.

\paragraph{Is the meta-level gain only its second candidate?} A registered control, \emph{numeric}, trains the same two candidates per round and searches the same edit space without the LLM (\cref{tab:budget}). On Goal it changes Official success by \Suppnumfix{} over \texttt{fixed}, against \Suppllmfix{} for the LLM. Both are program adaptation within the same gated loop; \cref{tab:supp-changes} shows what the LLM changed and why.

\paragraph{What does the LLM cost?} The \SuppLLMCalls{} logged calls of the eight self-repair runs cost \SuppLLMUSD{} USD together, at most \SuppLLMMaxUSD{} USD per run (\cref{tab:llmcost}).

\paragraph{Can test scenes reach training or selection?} No. Every placement set is drawn from its own seed range (\cref{sec:impl}); training, the gate, the audit and the meta level never see Official or Fresh outcomes. Each comparison of \cref{tab:components} and its analysis were registered before its evaluations existed, except one descriptive pair, the Goal fresh-scene meta-level effect (\cref{tab:claims}).

\paragraph{How was the RL baseline tuned, and what does each cost include?} RL's recipe was chosen by \SuppPilotRuns{} pilot runs (\cref{tab:pilots}; \SuppPilotGPUh{} GPU-hours), and RL-36 doubles its training (\cref{sec:rl}). The loop's GPU-hours include all of its candidates and its gate and audit evaluations (\cref{sec:impl}).

\paragraph{Why is one seed's continuation not reported?} Its practice data collapsed while its validation score stayed level. The exclusion was decided post hoc and recorded with its reason in a dated addendum before the paper was finalised; that seed's first six rounds are reported (\cref{sec:costsupp}).

\paragraph{Can the numbers be reproduced?} The code release's \texttt{README.md} maps each table and figure of the paper to its script and records, and \texttt{scripts/supplement/} regenerates every table of this supplement (\cref{sec:code}).

\section{Implementation details}
\label{sec:impl}

\paragraph{Policy, rollouts and updates.}
The policy is OpenVLA-OFT (7B) with parallel decoding of 8-action chunks (56 discrete tokens over 256 bins), started from the SimpleVLA-RL supervised checkpoint trained on one demonstration per task. Each candidate trains a LoRA adapter (rank 32, $\alpha{=}16$, no dropout, on the attention projections) with AdamW (no weight decay), gradient-norm clipping at 1.0 and bf16 autocast, micro-batches of 16. Rollouts use temperature $T$ for practice and greedy decoding for gate, audit and test episodes, with a limit of 512 environment steps. GRPO uses the clipped policy-ratio objective with the group-centred advantage within each instance; with \texttt{matched} ratios the old log-probabilities are computed on exactly the micro-batches that are trained, and \texttt{episode} aggregation gives every episode the same total weight. BC maximises the likelihood of the kept successful attempts. Within a round, each of the three blocks continues from the previous block's adapter with a fresh optimiser. \Cref{tab:recipes} lists the two starting programs.

\begin{table}[t]
\centering\footnotesize
\setlength{\tabcolsep}{3pt}
\caption{\textbf{Starting programs} (identical on both suites, read from the retained records of the \texttt{fixed} runs). GRPO: the working recipe of the recursion and gate-strictness studies. BC: the self-training recipe that degrades without a gate, the start of the gate-removal and self-repair studies. Values in parentheses are carried by the BC program and take effect only under GRPO, \eg after the meta level switches the operator (\cref{tab:supp-changes}).}
\label{tab:recipes}
\resizebox{\columnwidth}{!}{%
\begin{tabular}{@{}lll@{}}
\toprule
Setting (program field) & GRPO & BC \\
\midrule
Instances (scenes) per block {\tiny(\texttt{curriculum.cases})} & 64 & 64 \\
Allocation over tasks {\tiny(\texttt{curriculum.allocation})} & uniform & uniform \\
Attempts per instance $K$ {\tiny(\texttt{solver.samples})} & 8 & 8 \\
Sampling temperature $T$ {\tiny(\texttt{solver.temperature})} & 1.6 & 1.6 \\
Data rule {\tiny(\texttt{curation.rule})} & mixed\_groups & verified \\
Success band of kept groups {\tiny(\texttt{curation.band})} & $[0.0, 1.0]$ & -- \\
Data budget per block {\tiny(\texttt{curation.budget})} & 400 & 400 \\
Successful attempts kept per instance {\tiny(\texttt{curation.per\_case})} & -- & 2 \\
Update operator {\tiny(\texttt{update.operator})} & grpo & bc \\
Learning rate {\tiny(\texttt{update.lr})} & $3\times10^{-5}$ & $1\times10^{-5}$ \\
Optimiser steps per block {\tiny(\texttt{update.steps})} & 100 & 100 \\
Batch size {\tiny(\texttt{update.batch})} & 64 & 64 \\
Advantage {\tiny(\texttt{curation.adv})} & center & (center) \\
Policy-ratio clip (low, high) {\tiny(\texttt{update.clip\_*})} & 0.2, 0.28 & (0.2, 0.28) \\
Policy ratio {\tiny(\texttt{update.ratio})} & matched & (matched) \\
Loss aggregation {\tiny(\texttt{update.loss\_agg})} & episode & (episode) \\
BC likelihood temperature {\tiny(\texttt{update.bc\_temp})} & -- & unit \\
\midrule
\multicolumn{3}{@{}l}{\emph{Shared by every run}} \\
LoRA rank / $\alpha$ / dropout & \multicolumn{2}{l}{32 / 16 / 0} \\
LoRA targets & \multicolumn{2}{l}{attention q, k, v, o projections} \\
Optimiser & \multicolumn{2}{l}{AdamW, weight decay 0, grad-norm clip 1.0, bf16} \\
Micro-batch & \multicolumn{2}{l}{16} \\
Blocks per round / rounds & \multicolumn{2}{l}{3 / 12 (GRPO) or 6 (BC)} \\
Gate / audit set & \multicolumn{2}{l}{20 / 20 placements per task (200 episodes each), greedy} \\
Episode limit / action chunk & \multicolumn{2}{l}{512 environment steps / 8 actions} \\
\bottomrule
\end{tabular}}
\end{table}

\paragraph{Evaluation sets.}
All placements are drawn by LIBERO's own sampler from disjoint seed ranges, so no placement is shared between uses. Practice instances use a range indexed by run seed, round, task and instance; the per-round gate sets (20 placements per task, 200 episodes, resampled every round) and the audit set (20 per task) use separate ranges; the Fresh test uses 1{,}000 placements (100 per task) from a range used by nothing else, identical for every run of a suite. Official uses the benchmark's 50 fixed initial states per task. Neither test is seen by training, gating, the meta level or the audit. The initial policy's reference on each test is the per-episode mean of its repeated evaluations on the same platform (four Fresh evaluations per suite and platform), and every paired comparison is formed within a platform.

\paragraph{Gate measurement noise.}
A gate score is a success rate over 200 greedy episodes, so at a success rate near 60\% its binomial standard error is about 3.5 points. Under an unchanged training program, the difference between a candidate and its parent on the same round's gate set has a standard deviation of 3.4 points in the run records. Differences of a few points are within this noise; the drops the gate refuses in ungated BC training are larger (\cref{sec:rq3}), and the relaxed \texttt{guard} gate that tolerates 5-point dips gives the same final results within the CIs (\cref{tab:perseed}, gate-strictness rows). The \texttt{llm\_harness} proposer is told this noise level in its prompt.

\paragraph{Compute and GPU-hours.}
The studies run on NVIDIA A800 80\,GB GPUs on two platforms: a cloud GPU pool (platform Y; seeds 1--3) and a dedicated server (platform A; seed 0 where a seed-0 run exists). Some RL pilots ran on a second cloud pool (platform B). Most runs occupy one GPU; some meta-level runs used two or three. GPU-hours are reserved GPU-hours: GPUs requested times job time, summed over a run's job sessions; this upper-bounds the GPU time actually used. A run's job time contains its practice rollouts, training, gate and audit evaluations and meta-level proposals. LLM inference is excluded. The tuned-RL total of \cref{sec:rl} adds the 213.4 GPU-hours of the 21 pilot runs that chose its recipe (\cref{tab:pilots}) to its own training.

\section{Meta level}
\label{sec:metasupp}
\paragraph{Proposer and edit space per study.} The self-repair runs on LIBERO-Goal use the plain proposer of \cref{sec:meta} (\texttt{llm}: one call per round, the records of earlier candidates in the prompt) over the harness space, without compute cap. The self-repair runs on LIBERO-Spatial use the workspace proposer (\texttt{llm\_harness}: several turns of listing, reading and searching files) over the harness space with per-task allocation, under the compute cap, with numeric search as the fallback for invalid proposals. Both call the same model. In the released records the Goal runs carry the label \texttt{llm\_harness}, after their edit space. \Cref{tab:supp-space} lists the edit space, \cref{fig:supp-prompt} the prompts; two first proposals are quoted verbatim below, and \cref{tab:supp-changes} every accepted change.

\IfFileExists{supp/meta_space.tex}{\begin{table*}[t]
\centering
\scriptsize
\caption{\textbf{Edit space of the meta level.} The \emph{knob} space has ten fields; the \emph{harness} space adds nine procedural fields. Ranges are inclusive; lists are the allowed choices. The descriptions are the neutral field descriptions given to the proposer. Proposals outside the space are rejected or reduced to their allowed fields, and invalid proposals fall back to numeric search.}
\label{tab:supp-space}
\setlength{\tabcolsep}{3pt}
\renewcommand{\arraystretch}{1.08}
\begin{tabular*}{\textwidth}{@{\extracolsep{\fill}}rlp{0.21\textwidth}p{0.53\textwidth}@{}}
\toprule
& field & range / choices & what it changes (as described to the proposer) \\
\midrule
\multicolumn{4}{@{}l}{\textbf{Knob space (10 fields)}} \\
1 & \texttt{curriculum.cases} & [40, 400] & scenes practised per update step \\
2 & \texttt{curriculum.allocation} & uniform, progress & 'uniform' spreads scenes evenly over tasks; 'progress' gives more scenes to tasks whose recent tries had both outcomes \\
3 & \texttt{solver.samples} & [4, 16] & tries per scene \\
4 & \texttt{solver.temperature} & [1, 2] & sampling temperature of the tries \\
5 & \texttt{curation.rule} & mixed\_groups, verified & 'mixed\_groups' keeps every try of scenes whose success fraction lies strictly inside curation.band (used by grpo); 'verified' keeps successful tries only (used by bc) \\
6 & \texttt{curation.budget} & [40, 1000] & at most this many scenes (grpo) or tries (bc) per step \\
7 & \texttt{update.operator} & grpo, bc & 'grpo' (clipped policy gradient with those advantages) or 'bc' (maximise the likelihood of the kept tries) \\
8 & \texttt{update.lr} & [2e-6, 1e-4] & learning rate of one update \\
9 & \texttt{update.steps} & [50, 1500] & optimiser steps of one update \\
10 & \texttt{update.batch} & [16, 128] & batch size of one update \\
\addlinespace[2pt]
\multicolumn{4}{@{}l}{\textbf{Harness space: the knob fields plus nine procedural fields}} \\
1 & \texttt{curation.band} & [0, 1], [0.1, 0.9], [0.2, 0.8], [0.3, 0.7] & [lo, hi] success-fraction band of the scenes that are used \\
2 & \texttt{curation.per\_case} & [1, 8] & at most this many successful tries per scene (bc) \\
3 & \texttt{curriculum.floor} & [0, 0.3] & minimum share of a task under 'progress' \\
4 & \texttt{update.clip\_low} & [0.1, 0.4] & lower end of the clip range of the policy ratio in grpo \\
5 & \texttt{update.clip\_high} & [0.1, 0.5] & upper end of the clip range of the policy ratio in grpo \\
6 & \texttt{update.ratio} & legacy, matched & the 'old' log-probabilities of the ratio are computed in a separate pass over all items ('legacy') or on exactly the micro-batches that are trained ('matched') \\
7 & \texttt{update.loss\_agg} & token, episode & the loss averages over tokens ('token') or gives every episode the same total weight ('episode') \\
8 & \texttt{update.bc\_temp} & unit, rollout & bc's likelihood uses logits at temperature 1 ('unit') or divided by the sampling temperature of the try ('rollout') \\
9 & \texttt{curation.adv} & center, zscore & the per-try advantage in a scene is reward minus the scene mean ('center') or that divided by the scene's standard deviation ('zscore') \\
\addlinespace[2pt]
\multicolumn{4}{@{}l}{\textbf{Spatial runs only: per-task allocation (space v2)}} \\
-- & \texttt{curriculum.cases} & [16, 400] & lower bound widened from 40 so a capped proposal can trade scenes for tries \\
-- & \texttt{curriculum.allocation} & uniform, progress, weights & adds 'weights': scenes split over tasks in proportion to curriculum.task\_weights \\
-- & \texttt{update.steps} & [25, 1500] & lower bound widened from 50 so a capped proposal can trade steps for batch size \\
-- & \texttt{curriculum.task\_weights} & one weight per task in [0, 4] & one weight per task, in task order; with allocation 'weights' the scenes are split over tasks in proportion to it (0 gives a task no scenes) \\
\bottomrule
\end{tabular*}
\vspace{3pt}

\parbox{\textwidth}{\scriptsize
\textbf{Coupling.} \texttt{update.operator} and \texttt{curation.rule} move together: \texttt{grpo} with \texttt{mixed\_groups}, \texttt{bc} with \texttt{verified}. \textbf{Fields absent from a program} (\texttt{update.ratio}, \texttt{update.loss\_agg}, \texttt{update.bc\_temp}, \texttt{curation.adv}) keep the run's configured behaviour. \textbf{Compute cap} (Spatial runs): a procedure may not do more work per update step than the starting procedure. curriculum.cases x solver.samples must stay at or below 512 tries, and update.steps x update.batch at or below 6400 examples. A procedure over the cap is shrunk proportionally (scenes and tries by the same factor; steps and batch by the same factor), so choose how to split the budget yourself.
\textbf{Space used by each arm.} Goal self-repair runs (seeds 0-3): single-call proposer over the harness space, no compute cap. Spatial self-repair runs (seeds 0-3): workspace proposer over the harness space plus per-task allocation, with the compute cap (512 tries and 6400 training examples per block).}
\end{table*}
}{}
\IfFileExists{supp/meta_prompt.tex}{\begin{figure*}[p]
\centering
\begin{minipage}{\textwidth}\scriptsize
\textbf{(a) Single-call proposer (LIBERO-Goal self-repair runs).} One call per round; the reply is parsed as JSON. Slots in angle brackets are filled each round: \texttt{<program>} with the current program, \texttt{<space>} and \texttt{<space\_doc>} with the edit space and its field descriptions (\cref{tab:supp-space}), \texttt{<history>} with the records of earlier candidates (program changes, validation scores and their change against the parent, acceptance, per-block sampled success, share of mixed-outcome scenes, item counts), \texttt{<k>} with the number of proposals.
\begin{verbatim}
You tune the training procedure of a robot policy that improves itself in simulation.

Each round, the current policy practises on fresh random scenes: curriculum.cases scenes, solver.samples tries
each, sampled at solver.temperature. A simulator marks every try success or failure, and that is the only
feedback. Curation picks the training items: "mixed_groups" (scenes with both outcomes; used with operator "grpo")
or "verified" (successful tries; used with operator "bc"). The update then trains the policy (operator, lr, steps,
batch). A round runs this practise-and-update loop a few times, each time with the newest policy. A gate score,
the greedy success rate on unseen scenes, decides whether a round's new policy is kept and passed on.

Current procedure (JSON):
<program>

You may change ONLY these fields (numbers are [min, max]; lists are the allowed choices; grpo needs mixed_groups,
bc needs verified). Copy every other field of the procedure unchanged; changes to other fields are discarded:
<space>
<space_doc>

History, oldest first. For each candidate: the fields that differed from its parent's procedure, its gate score
and the change against its parent, whether it was kept, and training-side signals per inner step (sampled success
rate, share of scenes with both outcomes, number of training items) plus the training loss at the first and last
update step:
<history>

First diagnose in one or two sentences what limits improvement, for example too few scenes with both outcomes,
sampled success collapsing, a loss that keeps rising, or a gate that does not move. Then propose <k> DIFFERENT
procedures to try next. Change only allowed fields. Answer with JSON only:
{"diagnosis": "...", "proposals": [{"why": "<one sentence>", "procedure": <complete procedure JSON>}, ...]}
\end{verbatim}
\textbf{(b) Workspace proposer (LIBERO-Spatial self-repair runs).} Up to six turns per round; each turn either calls tools over the workspace or gives the final answer. Only the parts that differ from (a) are shown.
\begin{verbatim}
You are the meta level of a robot policy that improves itself in simulation: you choose the training procedure of
the next round. Your goal is the success rate of the FINAL policy on unseen scenes after the last round.

[... the description of a round, as in the Goal prompt above ...]

Each round trains <children> children of the current policy on the same scenes: c0 with the current procedure and
the others with your proposals. All are scored on a new gate set together with the parent; the best child is kept
if it scores at least the parent, and its procedure becomes the current one. This is round <round> of <rounds>.

You work in a workspace of files (index below). Reply with exactly ONE JSON object per turn and nothing else:
tools are called by writing that JSON in your reply; there is no other tool interface and no tag syntax. Either
tool calls, at most <max_calls_per_turn> per turn:
  {"calls": [{"tool": "read", "path": "run/nodes/r01c1.json"}, {"tool": "grep", "pattern": "bc", "path":
  "memory/prior"},
             {"tool": "ls", "path": "memory/prior"}]}
("read" takes optional "start" (first line, from 1) and "lines"; "grep" takes a regular expression and a file or a
folder), or your final answer:
  {"diagnosis": "<two or three sentences: what limits improvement, and the evidence>",
    "proposals": [{"why": "<one sentence>", "procedure": <complete procedure JSON>}, ...],
    "note": "<what you tried this round, what the evidence so far shows, what to try next>"}
Give <k> DIFFERENT proposals. "note" is appended to memory/notes.md, which you will find again in later rounds.
You have <turns_left> turn(s) left including this one; on your last turn you must give the final answer.

[... current procedure, allowed fields and their descriptions, as in the Goal prompt ...]

How to read the evidence.
* A gate is the greedy success rate on unseen scenes (<gate_per_task> per task). Re-scoring one policy on another
  gate set moves the score by about 3 points (one standard deviation); a child against its parent on the same set
  moves by about 3.4 points even when the procedure is unchanged. One gate change under about 6 points is not
  evidence.
* Evidence that survives the noise: the same direction in several rounds or several runs (run/ledger.tsv,
  memory/prior), a per-task pattern that repeats, or a training-side signal with many tries behind it (sampled
  success over hundreds of tries per inner step; a loss that rises during an update; scenes with both outcomes
  disappearing).
* A change that is kept is not thereby proven: the better of two noisy children is kept about half the time by
  luck.

Workspace index (path, lines, what it is):
<index>
\end{verbatim}
\end{minipage}
\caption{\textbf{Proposer prompts} (verbatim templates from the released code). Both proposers use the same model; the history, the workspace (\cref{tab:supp-workspace}) and the memory never contain audit or test results.}
\label{fig:supp-prompt}
\end{figure*}

\begin{table}[t]
\centering
\scriptsize
\caption{\textbf{Workspace of the workspace proposer} as indexed in its prompt (lines: size at the start of a run). One \texttt{run/nodes/<node>.json} file per candidate is added as candidates finish; \texttt{memory/notes.md} accumulates the proposer's own notes. The memory was frozen before launch.}
\label{tab:supp-workspace}
\setlength{\tabcolsep}{3pt}
\begin{tabular}{@{}p{0.43\linewidth}rp{0.43\linewidth}@{}}
\toprule
file & lines & content \\
\midrule
\texttt{current/procedure.json} & 1 & \raggedright the current procedure \tabularnewline
\texttt{memory/README.md} & 15 & \raggedright what the memory holds and how its tables are laid out \tabularnewline
\texttt{memory/notes.md} & 1 & \raggedright your own notes from earlier rounds of this run \tabularnewline
\texttt{memory/playbook.md} & 86 & \raggedright lessons distilled from finished runs before this run started; read it first \tabularnewline
\texttt{memory/prior/INDEX.tsv} & 33 & \raggedright the finished runs behind the playbook: suite, start procedure, outcome \tabularnewline
\texttt{run/ledger.tsv} & 1 & \raggedright this run's gate changes pooled per edited field=value \tabularnewline
\texttt{run/overview.tsv} & 1 & \raggedright every candidate of THIS run, one line each \tabularnewline
\texttt{run/tasks.tsv} & 1 & \raggedright per task: parent gate, child gate and sampled success per inner step \tabularnewline
\texttt{memory/prior/<run>.tsv} & 32 files & \raggedright one table per finished run, same columns as run/overview.tsv; the runs are listed in memory/prior/INDEX.tsv \tabularnewline
\bottomrule
\end{tabular}
\end{table}
}{}
\IfFileExists{supp/meta_example.tex}{\paragraph{Example proposer replies.} The replies below are quoted from the run logs (whitespace normalised; [...] marks a cut). In both runs the model's first proposal replaces BC on successful tries with GRPO on mixed-outcome scenes.

\noindent\textbf{LIBERO-Goal, seed 1 (platform Y), round 1, single-call proposer.}
\begin{quote}\footnotesize
\textit{Diagnosis:} ``Only the seed exists (gate 0.61) and there are no inner-loop signals, so the limiting factor is uncertain. The setup looks conservative: BC on verified tries at temperature 1.6 and lr 1e-5 learns from noisy high-temperature samples and gives no contrast signal from failures. bc\_temp 'unit' also mismatches the 1.6 sampling temperature.''\\
\textit{Why (proposal c1):} ``Switch to GRPO on mixed-outcome scenes, which uses failures as contrast, with a lower sampling temperature for cleaner samples, more tries per scene, progress allocation toward informative tasks, and a somewhat higher lr.''\\
\textit{Program change after validation:} operator bc$\to$grpo; rule verified$\to$mixed\_groups; advantage center$\to$zscore; allocation uniform$\to$progress; scenes 64$\to$128; T 1.6$\to$1.2; lr 1e-5$\to$2e-5.\\
\textit{Outcome:} Round 1 validation success: r01c0 57.0\%, r01c1 59.5\%; retained: r01c1 (the proposal).
\end{quote}
{\scriptsize Proposed procedure as written by the model:
\begin{verbatim}
{"curation": {"adv": "zscore", "band": [0.0, 1.0],
  "budget": 400, "per_case": 2, "rule": "mixed_groups"
  }, "curriculum": {"allocation": "progress",
  "cases": 128, "floor": 0.05}, "solver": {"greedy":
  0, "samples": 8, "temperature": 1.2}, "update":
  {"batch": 64, "bc_temp": "unit", "clip_high": 0.28,
  "clip_low": 0.2, "loss_agg": "episode", "lr":
  2e-05, "operator": "grpo", "ratio": "matched",
  "steps": 100}}
\end{verbatim}}

\noindent\textbf{LIBERO-Spatial, seed 1 (platform Y), round 1, workspace proposer.} Tool calls before the answer (4): read memory/playbook.md; read run/nodes/n0.json; read run/tasks.tsv; read memory/README.md.
\begin{quote}\footnotesize
\textit{Diagnosis:} ``The start procedure is BC at lr 1e-5, which the playbook marks as unreliable: every collapse in the prior runs was a BC run, and BC gave no gain on libero\_goal. I have no practice curves from this run yet (n0 is only the seed, gate 0.595), so this rests on the prior runs, not on evidence from this run.''\\
\textit{Why (proposal c1):} ``Switch to GRPO on mixed-group scenes with lr 1e-4, matched ratio and token loss, the setup that kept 5 of 6 rounds in all four replicas (sums of kept changes +0.135 to +0.180).''\\
\textit{Program change after validation:} operator bc$\to$grpo; rule verified$\to$mixed\_groups; loss episode$\to$token; lr 1e-5$\to$1e-4.\\
\textit{Outcome:} Round 1 validation success: r01c0 59.5\%, r01c1 62.0\%; retained: r01c1 (the proposal).
\end{quote}
}{}
\IfFileExists{supp/meta_changes.tex}{\begin{table*}[t]
\centering
\scriptsize
\caption{\textbf{Program changes accepted in the self-repair runs} (six rounds from the BC recipe, two candidates a round). Each row is a round whose retained candidate carries a program different from its parent's; the change is the field-by-field difference between the two programs. Gate: validation success of the retained candidate and of the other candidate on that round's validation set (\%). Kept: rounds that retained a candidate. Final: success of the retained version on the Official test (three 500-episode evaluations) and on 1{,}000 Fresh placements, with the \texttt{fixed} arm of the same seed in parentheses; --: not evaluated. Platform Y: cloud GPU pool; platform A: A800 server.}
\label{tab:supp-changes}
\setlength{\tabcolsep}{3pt}
\renewcommand{\arraystretch}{1.1}
\begin{tabular*}{\textwidth}{@{\extracolsep{\fill}}llrlp{0.47\textwidth}l@{}}
\toprule
run & kept & round & gate & accepted program change & final Official / Fresh \\
\midrule
\multicolumn{6}{@{}l}{\textbf{LIBERO-Goal}} \\
seed 0 (A) & 1,2,3,4,5,6 & R1 & 59.5 / 52.5 & operator bc$\to$grpo; rule verified$\to$mixed\_groups; allocation uniform$\to$progress; scenes 64$\to$128; T 1.6$\to$1.2; lr 1e-5$\to$2e-5; steps 100$\to$200 & 60.9 (57.4) / -- \\
 &  & R2 & 65.5 / 60.5 & advantage center$\to$zscore; band [0, 1]$\to$[0.1, 0.9]; budget 400$\to$600; scenes 128$\to$192; floor 0.05$\to$0.1; tries 8$\to$12; lr 2e-5$\to$4e-5; steps 200$\to$300 &  \\
 &  & R3 & 62.0 / 61.0 & budget 600$\to$800; scenes 192$\to$256; batch 64$\to$96; lr 4e-5$\to$5e-5; steps 300$\to$400 &  \\
 &  & R5 & 69.5 / 67.0 & budget 800$\to$1000; scenes 256$\to$320; batch 96$\to$128; lr 5e-5$\to$3e-5 &  \\
\addlinespace[2pt]
seed 1 (Y) & 1,2,3,4,5 & R1 & 59.5 / 57.0 & operator bc$\to$grpo; rule verified$\to$mixed\_groups; advantage center$\to$zscore; allocation uniform$\to$progress; scenes 64$\to$128; T 1.6$\to$1.2; lr 1e-5$\to$2e-5 & 60.1 (56.1) / 64.2 (57.5) \\
 &  & R2 & 65.5 / 61.5 & budget 400$\to$600; tries 8$\to$12; lr 2e-5$\to$5e-5; steps 100$\to$200 &  \\
 &  & R5 & 65.5 / 61.5 & band [0, 1]$\to$[0.1, 0.9]; batch 64$\to$128 &  \\
\addlinespace[2pt]
seed 2 (Y) & 1,2,3,4,5,6 & R1 & 64.5 / 61.0 & operator bc$\to$grpo; rule verified$\to$mixed\_groups; budget 400$\to$200; allocation uniform$\to$progress; scenes 64$\to$128; T 1.6$\to$1.2; lr 1e-5$\to$2e-5; steps 100$\to$200 & 61.3 (58.7) / 64.5 (61.0) \\
 &  & R2 & 64.0 / 61.0 & advantage center$\to$zscore; tries 8$\to$12; lr 2e-5$\to$4e-5; steps 200$\to$300 &  \\
 &  & R3 & 65.5 / 62.5 & band [0, 1]$\to$[0.1, 0.9]; budget 200$\to$300; scenes 128$\to$192; batch 64$\to$128; lr 4e-5$\to$3e-5 &  \\
 &  & R4 & 63.5 / 61.5 & T 1.2$\to$1; lr 3e-5$\to$2e-5; steps 300$\to$400 &  \\
\addlinespace[2pt]
seed 3 (Y) & 1,3,4,5,6 & R4 & 62.5 / 61.0 & allocation uniform$\to$progress; scenes 64$\to$200; floor 0.05$\to$0.1; T 1.6$\to$1.3; lr 1e-5$\to$2e-5; steps 100$\to$200 & 57.9 (57.5) / 61.9 (59.1) \\
 &  & R6 & 64.5 / 59.0 & band [0, 1]$\to$[0.1, 0.9]; budget 400$\to$600; tries 8$\to$16 &  \\
\addlinespace[2pt]
\multicolumn{6}{@{}l}{\textbf{LIBERO-Spatial}} \\
seed 0 (A) & 1,2,3,4,5,6 & R1 & 58.0 / 56.5 & operator bc$\to$grpo; rule verified$\to$mixed\_groups; loss episode$\to$token; lr 1e-5$\to$1e-4 & 61.8 (55.1) / 69.0 (64.3) \\
\addlinespace[2pt]
seed 1 (Y) & 1,2,3,4,5,6 & R1 & 62.0 / 59.5 & operator bc$\to$grpo; rule verified$\to$mixed\_groups; loss episode$\to$token; lr 1e-5$\to$1e-4 & 61.4 (58.8) / 66.9 (58.9) \\
 &  & R6 & 69.5 / 68.5 & steps 100$\to$50 (numeric fallback) &  \\
\addlinespace[2pt]
seed 2 (Y) & 1,2,3,5 & R3 & 65.0 / 63.0 & allocation uniform$\to$weights; task weights (2,0,0,0,2,2,0,2,2,2) & 54.2 (59.0) / 63.7 (63.9) \\
 &  & R5 & 68.0 / 62.5 & lr 1e-5$\to$3e-5 &  \\
\addlinespace[2pt]
seed 3 (Y) & 1,3,4,6 & R3 & 63.5 / 60.5 & allocation uniform$\to$weights; task weights (3,1.5,0.3,0.3,1,1.5,0.2,3,1.5,0.5) & 54.6 (56.9) / 61.4 (59.5) \\
 &  & R4 & 60.5 / 57.5 & per scene 2$\to$4 &  \\
\bottomrule
\end{tabular*}
\end{table*}
}{}

\paragraph{The same candidate budget without the LLM.} The fixed-program arms train one candidate per round and the meta-level arms two (\cref{tab:controls}). The registered control \emph{numeric} isolates the second candidate: it proposes by numeric search over the same edit space, with two candidates per round like the LLM arm (\cref{tab:budget}).

\begin{table}[t]
\centering\footnotesize
\setlength{\tabcolsep}{3pt}
\caption{\textbf{The same candidate budget without the LLM.} The registered control arm \emph{numeric} searches the same edit space with the same two candidates per round as the LLM arm. Differences (pp) and 95\% CIs as in \cref{tab:perseed}. Goal: seeds 0--3, Official only (the numeric arm has no Fresh evaluation). Spatial: seed 0 (platform A), the only seed with a numeric run.}
\label{tab:budget}
\resizebox{\columnwidth}{!}{%
\begin{tabular}{@{}lccc@{}}
\toprule
 & Goal & \multicolumn{2}{c}{Spatial, seed 0} \\
\cmidrule(lr){2-2}\cmidrule(lr){3-4}
Comparison & Official & Official & Fresh \\
\midrule
numeric $-$ \texttt{fixed} & $+1.50$ {\tiny $[-0.33, +3.38]$} & $-1.27$ {\tiny $[-4.93, +2.53]$} & $+6.70$ {\tiny $[+4.00, +9.40]$} \\
LLM $-$ \texttt{fixed} & $+2.65$ {\tiny $[+0.73, +4.58]$} & $+6.67$ & $+4.70$ {\tiny $[+2.60, +6.80]$} \\
LLM $-$ numeric & $+1.15$ {\tiny $[-0.63, +2.92]$} & $+7.93$ {\tiny $[+4.40, +11.47]$} & $-2.00$ {\tiny $[-4.50, +0.60]$} \\
\bottomrule
\end{tabular}}
\\[2pt]{\footnotesize Spatial seed 0, LLM $-$ \texttt{fixed}, Official: the per-seed value of \cref{tab:perseed} (three evaluations per run; no single-seed CI was registered).}
\end{table}

\paragraph{LLM usage.} \Cref{tab:llmcost} lists every logged call of the eight self-repair runs. The LLM's cost is reported separately from the GPU-hours.

\begin{table}[t]
\centering\footnotesize
\setlength{\tabcolsep}{4pt}
\caption{\textbf{LLM usage of the meta level.} Every logged call of the six-round self-repair runs of \mref{tab:components} (Claude Sonnet 5.5): logged calls, tokens and the cost the API reported with each call; \emph{LLM rounds}: rounds whose new candidate the LLM wrote, from the round records (six rounds per run). Where the two counts differ, a call was retried or not logged. Spatial runs use several turns of the workspace proposer per call. These costs are not part of the GPU-hours.}
\label{tab:llmcost}
\resizebox{\columnwidth}{!}{%
\begin{tabular}{@{}lcrrrrr@{}}
\toprule
Suite & Seed & LLM rounds & Calls & Input tokens & Output tokens & Cost (USD) \\
\midrule
Goal & 0 (A) & 6 & 7 & 25{,}441 & 3{,}746 & 0.09 \\
Goal & 1 & 6 & 7 & 25{,}864 & 3{,}203 & 0.08 \\
Goal & 2 & 6 & 5 & 15{,}999 & 2{,}204 & 0.05 \\
Goal & 3 & 6 & 6 & 21{,}332 & 3{,}781 & 0.08 \\
Spatial & 0 (A) & 6 & 7 & 157{,}129 & 33{,}855 & 0.65 \\
Spatial & 1 & 5 & 6 & 115{,}933 & 19{,}544 & 0.43 \\
Spatial & 2 & 6 & 6 & 118{,}512 & 20{,}359 & 0.44 \\
Spatial & 3 & 6 & 6 & 165{,}203 & 37{,}525 & 0.71 \\
\midrule
All 8 runs & & 47 & 50 & 645{,}413 & 124{,}217 & 2.53 \\
\bottomrule
\end{tabular}}
\end{table}

\section{Additional results}
\label{sec:results_supp}

\subsection{Per-seed values}
\Cref{tab:perseed} repeats every row of \cref{tab:components} with the success of each run's retained policy. The effects are computed per episode and seed-averaged, so a row's mean is the bootstrap estimate reported in the paper. Recursion and the gate are positive in every seed on both suites and both tests (\SuppRecPos{} of \SuppRecTot{} and \SuppGatePos{} of \SuppGateTot{} seed-level differences).

\begin{table*}[t]
\centering\footnotesize
\setlength{\tabcolsep}{4pt}
\caption{\textbf{Per-seed results behind \mref{tab:components}.} Success (\%) of the policy each run retained at its end, for the arm and its control, and their difference (pp). Official: 500 fixed benchmark states, mean of three evaluations per run. Fresh: 1{,}000 sampled placements, one evaluation per run. Bottom line of each block: seed-averaged paired bootstrap mean and 95\% CI (10{,}000 draws), as in the paper. Seed 0 runs on platform A, seeds 1--3 on platform Y; pairs are formed within a platform. \mref{tab:gate} and \mref{fig:gates} show the single evaluation made during each run, so their values can differ slightly from these means.}
\label{tab:perseed}
\begin{tabular}{@{}lllcrrrcrrr@{}}
\toprule
 & & & & \multicolumn{3}{c}{Official} & & \multicolumn{3}{c}{Fresh} \\
\cmidrule(lr){5-7}\cmidrule(lr){9-11}
Suite & Comparison (setup) & Arm $-$ control & Seed & arm & control & $\Delta$ & & arm & control & $\Delta$ \\
\midrule
Goal & Recursion (GRPO/12) & \texttt{fixed} $-$ \texttt{nonrec} & 1 & 62.6 & 48.2 & $+14.4$ & & 60.9 & 52.7 & $+8.2$ \\
 & &  & 2 & 60.3 & 55.4 & $+4.9$ & & 66.9 & 59.5 & $+7.4$ \\
 & &  & 3 & 58.7 & 54.5 & $+4.2$ & & 62.9 & 56.6 & $+6.3$ \\
 & & & mean & \multicolumn{3}{r}{$+7.84$ {\tiny $[+5.22, +10.49]$}} & & \multicolumn{3}{r}{$+7.30$ {\tiny $[+5.47, +9.20]$}} \\
\midrule
Goal & Gate strictness (GRPO/12) & \texttt{guard} $-$ \texttt{fixed} & 1 & 60.9 & 62.6 & $-1.7$ & & 66.9 & 60.9 & $+6.0$ \\
 & &  & 2 & 62.3 & 60.3 & $+1.9$ & & 64.9 & 66.9 & $-2.0$ \\
 & &  & 3 & 61.1 & 58.7 & $+2.4$ & & 63.9 & 62.9 & $+1.0$ \\
 & & & mean & \multicolumn{3}{r}{$+0.89$ {\tiny $[-1.11, +2.89]$}} & & \multicolumn{3}{r}{$+1.67$ {\tiny $[+0.00, +3.30]$}} \\
\midrule
Goal & Gate (BC/6) & \texttt{fixed} $-$ \texttt{nogate} & 1 & 56.1 & 43.9 & $+12.2$ & & 57.5 & 40.3 & $+17.2$ \\
 & &  & 2 & 58.7 & 45.1 & $+13.6$ & & 61.0 & 47.8 & $+13.2$ \\
 & &  & 3 & 57.5 & 42.5 & $+15.0$ & & 59.1 & 43.1 & $+16.0$ \\
 & & & mean & \multicolumn{3}{r}{$+13.60$ {\tiny $[+10.53, +16.76]$}} & & \multicolumn{3}{r}{$+15.47$ {\tiny $[+13.20, +17.77]$}} \\
\midrule
Goal & LLM meta level (BC/6) & \texttt{llm} $-$ \texttt{fixed} & 0 (A) & 60.9 & 57.4 & $+3.5$ & & -- & -- & -- \\
 & &  & 1 & 60.1 & 56.1 & $+4.1$ & & 64.2 & 57.5 & $+6.7$ \\
 & &  & 2 & 61.3 & 58.7 & $+2.6$ & & 64.5 & 61.0 & $+3.5$ \\
 & &  & 3 & 57.9 & 57.5 & $+0.5$ & & 61.9 & 59.1 & $+2.8$ \\
 & & & mean & \multicolumn{3}{r}{$+2.65$ {\tiny $[+0.73, +4.58]$}} & & \multicolumn{3}{r}{$+4.33$ {\tiny $[+2.57, +6.10]$}} \\
\midrule\midrule
Spatial & Recursion (GRPO/12) & \texttt{fixed} $-$ \texttt{nonrec} & 1 & 60.2 & 49.8 & $+10.4$ & & 66.5 & 57.8 & $+8.7$ \\
 & &  & 2 & 63.7 & 53.3 & $+10.4$ & & 69.1 & 59.0 & $+10.1$ \\
 & &  & 3 & 59.2 & 54.0 & $+5.2$ & & 65.2 & 58.1 & $+7.1$ \\
 & & & mean & \multicolumn{3}{r}{$+8.67$ {\tiny $[+6.76, +10.69]$}} & & \multicolumn{3}{r}{$+8.63$ {\tiny $[+7.17, +10.10]$}} \\
\midrule
Spatial & Gate strictness (GRPO/12) & \texttt{guard} $-$ \texttt{fixed} & 1 & 62.3 & 60.2 & $+2.1$ & & 69.9 & 66.5 & $+3.4$ \\
 & &  & 2 & 61.1 & 63.7 & $-2.6$ & & 68.7 & 69.1 & $-0.4$ \\
 & &  & 3 & 60.5 & 59.2 & $+1.3$ & & 65.7 & 65.2 & $+0.5$ \\
 & & & mean & \multicolumn{3}{r}{$+0.29$ {\tiny $[-1.51, +2.13]$}} & & \multicolumn{3}{r}{$+1.17$ {\tiny $[-0.07, +2.40]$}} \\
\midrule
Spatial & Gate strictness (GRPO/6) & \texttt{guard} $-$ \texttt{fixed} & 0 (A) & 58.9 & 60.9 & $-1.9$ & & 67.0 & 65.9 & $+1.1$ \\
 & &  & 1 & 63.0 & 60.5 & $+2.5$ & & 64.9 & 64.3 & $+0.6$ \\
 & &  & 2 & 60.2 & 60.7 & $-0.5$ & & 65.4 & 65.0 & $+0.4$ \\
 & &  & 3 & 62.3 & 59.5 & $+2.9$ & & 64.5 & 65.2 & $-0.7$ \\
 & & & mean & \multicolumn{3}{r}{$+0.73$ {\tiny $[-0.77, +2.25]$}} & & \multicolumn{3}{r}{$+0.35$ {\tiny $[-0.62, +1.32]$}} \\
\midrule
Spatial & Gate (BC/6) & \texttt{fixed} $-$ \texttt{nogate} & 1 & 58.8 & 44.2 & $+14.6$ & & 58.9 & 50.9 & $+8.0$ \\
 & &  & 2 & 59.0 & 45.9 & $+13.1$ & & 63.9 & 53.2 & $+10.7$ \\
 & &  & 3 & 56.9 & 45.6 & $+11.3$ & & 59.5 & 53.1 & $+6.4$ \\
 & & & mean & \multicolumn{3}{r}{$+13.00$ {\tiny $[+10.44, +15.60]$}} & & \multicolumn{3}{r}{$+8.37$ {\tiny $[+6.67, +10.07]$}} \\
\midrule
Spatial & LLM meta level (BC/6) & \texttt{llm} $-$ \texttt{fixed} & 0 (A) & 61.8 & 55.1 & $+6.7$ & & 69.0 & 64.3 & $+4.7$ \\
 & &  & 1 & 61.4 & 58.8 & $+2.6$ & & 66.9 & 58.9 & $+8.0$ \\
 & &  & 2 & 54.2 & 59.0 & $-4.8$ & & 63.7 & 63.9 & $-0.2$ \\
 & &  & 3 & 54.6 & 56.9 & $-2.3$ & & 61.4 & 59.5 & $+1.9$ \\
 & & & mean & \multicolumn{3}{r}{$+0.53$ {\tiny $[-1.05, +2.13]$}} & & \multicolumn{3}{r}{$+3.60$ {\tiny $[+2.50, +4.72]$}} \\
\bottomrule
\end{tabular}
\end{table*}

\subsection{Registration}
Each comparison of \cref{tab:components}, its threshold and its analysis were registered in a dated addendum to the preregistration before the evaluations it compares existed, except the descriptive Goal fresh-scene meta-level pair (\cref{tab:claims}); the analysis was run once, by the registered script. The Fresh test was registered in one addendum (Nov.\ 10), before any run was evaluated on it, following the train-on-test analysis of \cref{sec:rq4}. The registered claims of the earlier six-round LIBERO-Goal study with a working GRPO program are in \cref{tab:goal}; there, at six rounds and on the official states, the loop and its controls differ by less than the registered bars, and the twelve-round extension of \cref{tab:recursion} was registered after it.

\begin{table*}[t]
\centering\footnotesize
\setlength{\tabcolsep}{4pt}
\caption{\textbf{Registration of the comparisons in \mref{tab:components}.} Each comparison, its threshold (bar: the difference must reach the bar with the 95\% CI above zero) and its analysis were fixed in a dated addendum to the preregistration before the evaluations it compares existed; dates are 2026 (month-day). The read is the date of the one-time analysis. Goal's fresh-scene meta-level effect is a descriptive pair of evaluations registered for other claims (no bar). The preregistration and all addenda are in the code release.}
\label{tab:claims}
\begin{tabular}{@{}llcccrcccr@{}}
\toprule
 & & \multicolumn{4}{c}{Official} & \multicolumn{4}{c}{Fresh} \\
\cmidrule(lr){3-6}\cmidrule(lr){7-10}
Comparison & Claim & registered & read & bar & result & registered & read & bar & result \\
\midrule
Goal, recursion (GRPO/12) & R3-long / FL3 & 11-07 & 11-08 & $+2$ & $+7.84$$^\checkmark$ & 11-10 & 11-10 & $+2$ & $+7.30$$^\checkmark$ \\
Goal, gate strictness (GRPO/12) & G2 & 11-10 & 11-12 & $+2$ & $+0.89$ & 11-10 & 11-12 & $+2$ & $+1.67$ \\
Goal, gate (BC/6) & NG1 & 11-10 & 11-12 & $+10$ & $+13.60$$^\checkmark$ & 11-10 & 11-12 & $+10$ & $+15.47$$^\checkmark$ \\
Goal, LLM meta level (BC/6) & S1 & 11-04 & 11-09 & $+3$ & $+2.65$ & -- & -- & -- & $+4.33$ \\
Spatial, recursion (GRPO/12) & R3-long-spatial / FL3 & 11-10 & 11-14 & $+2$ & $+8.67$$^\checkmark$ & 11-10 & 11-14 & $+2$ & $+8.63$$^\checkmark$ \\
Spatial, gate strictness (GRPO/12) & G2s & 11-12 & 11-14 & $+2$ & $+0.29$ & 11-12 & 11-14 & $+2$ & $+1.17$ \\
Spatial, gate strictness (GRPO/6) & GS2 & 11-11 & 11-12 & $+2$ & $+0.73$ & 11-11 & 11-12 & $+2$ & $+0.35$ \\
Spatial, gate (BC/6) & N1 & 11-10 & 11-12 & $+10$ & $+13.00$$^\checkmark$ & 11-10 & 11-12 & $+10$ & $+8.37$ \\
Spatial, LLM meta level (BC/6) & S1-spatial / FS1 & 11-10 & 11-14 & $+3$ & $+0.53$ & 11-10 & 11-14 & $+3$ & $+3.60$$^\checkmark$ \\
\bottomrule
\end{tabular}
\\[2pt]{\footnotesize $^\checkmark$ registered bar met. Gate strictness tests whether a relaxed gate (\texttt{guard}) improves on the strict gate by at least 2 points; it does not, and the component studies use the strict gate.}
\end{table*}

\begin{table*}[h]
  \centering
  \small
  \caption{\textbf{Registered claims on LIBERO-Goal} (official 500-episode test, 3-evaluation finals, pooled initial reference, seed-averaged paired bootstrap). Platform Y = cloud pool, A = A800. Arm names are run labels: \texttt{full} is numeric search; \texttt{llm} and \texttt{llm\_harness} are the plain LLM proposer over the knob and the harness space; \texttt{cap} adds the compute cap. Six rounds with the working GRPO program, seeds as listed.}
  \label{tab:goal}
  \begin{tabular}{@{}llrrcc@{}}
    \toprule
    claim & comparison & points & 95\% CI & bar & seeds \\
    \midrule
    R1 & \texttt{fixed} $-$ ref. (Y) & $+1.71$ & $[-0.76, +4.23]$ & $+5$ & 1--3 \\
    R1 & \texttt{full} $-$ ref. (Y) & $+1.28$ & $[-1.37, +3.99]$ & $+5$ & 1--3 \\
    R2 & \texttt{fixed} $-$ \texttt{shuffled} (Y) & $+1.22$ & $[-0.37, +2.85]$ & $+3$ & 0--3 \\
    R3 & \texttt{fixed} $-$ \texttt{nonrec} (Y) & $+1.72$ & $[-0.28, +3.68]$ & $+2$ & 0--3 \\
    R4 & \texttt{full} $-$ \texttt{fixed} (Y) & $-0.42$ & $[-2.40, +1.56]$ & $+2$ & 1--3 \\
    R4 & \texttt{llm\_harness} $-$ \texttt{fixed} (Y) & $-0.73$ & $[-2.58, +1.07]$ & $+2$ & 1--3 \\
    R4 & \texttt{llm} $-$ \texttt{fixed} (A) & $+1.04$ & $[-0.84, +3.00]$ & $+2$ & 1--3 \\
    R4 & \texttt{llm\_harness\_cap} $-$ \texttt{fixed} (A) & $+1.16$ & $[-0.76, +3.16]$ & $+2$ & 1--3 \\
    \midrule
    noise & half vs.\ half of the n0 evaluations (Y / A) & $+0.64$ / $+0.54$ & $[-0.07, +1.36]$ / $[-0.24, +1.35]$ & & \\
    \bottomrule
  \end{tabular}
\end{table*}

\subsection{Compute efficiency on LIBERO-Goal}
\label{sec:costsupp}
\paragraph{Per seed at the registered stop.} The continuation of \cref{sec:rl} keeps each run's adapted program fixed and continues with one candidate per round under the relaxed guard-rail gate until its registered stop: its planned last round, a job time limit set inside RL's GPU budget, or the registered cut-off date, whichever came first.

\begin{center}\footnotesize\setlength{\tabcolsep}{3pt}
\resizebox{\linewidth}{!}{%
\begin{tabular}{@{}lrrrrrrrr@{}}
\toprule
 & \multicolumn{2}{c}{RSI loop} & \multicolumn{3}{c}{tuned RL, 18 blocks} & \multicolumn{3}{c}{RL, 36 blocks} \\
\cmidrule(lr){2-3}\cmidrule(lr){4-6}\cmidrule(lr){7-9}
seed & GPU-h & fresh / official & GPU-h & fresh / official & $\Delta$ fresh & GPU-h & fresh / official & $\Delta$ fresh \\
\midrule
1 & 143 & 70.6 / 63.4 & 229 & 60.0 / 58.8 & $+10.6$ & 241 & 66.8 / 58.6 & $+3.8$ \\
2 & 238 & 66.7 / 62.5 & 230 & 62.2 / 59.0 & $+4.5$ & 239 & 65.5 / 60.9 & $+1.2$ \\
\midrule
mean & 190 & 68.7 / 62.9 & 230 & 61.1 / 58.9 & $+7.6$ & 240 & 66.2 / 59.7 & $+2.5$ \\
\bottomrule
\end{tabular}}
\end{center}
Success in \%; GPU-hours include RL's 213 pilot hours. The loop's version is the one each run retained when it stopped (addendum 35). The third registered seed's continuation is excluded (see below). Addendum 38 continues seed 1 up to RL's budget.
 Seed 0 on platform A (A800 server), continued the same way: against its tuned RL run (18 blocks), 67.6 vs 64.8 on the fresh scenes ($+2.8$ $[+0.5, +5.0]$) and 62.5 vs 62.7 on the official test, three evaluations per version.

\paragraph{At about the same total GPU-hours (addendum 38).}
\begin{center}\footnotesize\setlength{\tabcolsep}{4pt}
\resizebox{\linewidth}{!}{%
\begin{tabular}{@{}lrrrrr@{}}
\toprule
seed & RSI GPU-h & RSI fresh / official & RL-18 GPU-h & RL-18 fresh / official & $\Delta$ fresh \\
\midrule
1 & 241 & 72.9 / 64.9 & 229 & 60.0 / 58.8 & $+12.9$ \\
2 & 238 & 66.7 / 62.5 & 230 & 62.2 / 59.0 & $+4.5$ \\
\midrule
average & & & & & $+8.7$ $[+6.7, +10.8]$ \\
\bottomrule
\end{tabular}}
\end{center}
Each run continued with its program fixed up to about its RL budget; the version each run retained when it stopped. Fresh scenes, one evaluation per version (registered); seed-averaged paired bootstrap.

\paragraph{Repeated evaluations.} At the registered stop (first table above), the registered read of \cref{sec:rl} uses one fresh-scene evaluation of each final version (addendum 35). With three evaluations of each final version (addendum 37), the loop's difference is \SuppExtRLeighteenThree{} against RL-18 and \SuppExtRLthirtysixThree{} against RL-36.

\paragraph{Exclusion of the third seed's continuation.}
Practice success is the fraction of verified practice attempts in a round. For the third registered seed it fell from 58\% in round 6, the last round before the continuation, to 32\% in round 7, the first round of the continuation, then to 10\% in round 8 and 3\% by round 15, while its retained policy still scored 64.5\% on validation. Seeds 1 and 2 show no such drop: seed 1 went 66\%, 63\%, 67\% and seed 2 66--70\%, 59\%, 64\% over rounds 6--8. From round 8 on, that seed's candidates were trained on the collapsed practice data, and the gate refused all but one of them. The exclusion was decided post hoc on these records and recorded in a dated addendum before the paper was finalised; the first six rounds of that seed (\cref{sec:rq1}) and every other study's third seed are unaffected and reported.

\paragraph{RL recipe search.}
\Cref{tab:pilots} lists the pilot runs whose GPU-hours are counted in RL's total cost.

\begin{table}[t]
\centering\footnotesize
\setlength{\tabcolsep}{4pt}
\caption{\textbf{RL recipe search.} The 21 pilot runs that chose the tuned GRPO recipe before the main studies, with their recorded GPU-hours (18 timed; 213.4 GPU-hours on record, counted in RL's total cost in \mref{sec:rl} and \mref{fig:cost}). Run names state what each pilot varied (learning rate, optimiser steps, loss aggregation, operator, suite: \texttt{l10} = LIBERO-10). Platform B is a second cloud pool used only for pilots.}
\label{tab:pilots}
\resizebox{\columnwidth}{!}{%
\begin{tabular}{@{}llrr@{}}
\toprule
Pilot run & Platform & GPUs & GPU-hours \\
\midrule
\texttt{pilot2\_l10\_lr1e4} & B & 1 & 4.1 \\
\texttt{pilot2\_l10\_lr5e5} & B & 1 & 3.9 \\
\texttt{pilot3\_l10\_s1000} & B & 1 & 3.3 \\
\texttt{pilot3\_l10\_s400} & B & 1 & 2.5 \\
\texttt{pilot5\_curve} & B & 1 & 15.5 \\
\texttt{pilot7\_lr3e5} & B & 1 & 15.2 \\
\texttt{pilot7\_lr3e5\_ep} & B & 1 & 15.1 \\
\texttt{pilot\_l10\_bc} & B & 1 & 3.7 \\
\texttt{pilot\_l10\_grpo} & B & 1 & 4.1 \\
\texttt{pilot4\_short\_s1} & A & 1 & -- \\
\texttt{pilot6\_curve\_matched\_s1} & A & 1 & 24.1 \\
\texttt{pilot6\_curve\_matched\_s1.failed1} & A & 1 & -- \\
\texttt{pilot7\_lr3e5\_ep\_s1} & A & 1 & 22.5 \\
\texttt{pilot7\_lr3e5\_s1} & A & 1 & 22.1 \\
\texttt{oftg\_curve\_bc} & B & 1 & 11.5 \\
\texttt{oftg\_curve\_grpo} & B & 1 & 11.7 \\
\texttt{oftg\_curve\_grpo\_s2} & B & 1 & 12.0 \\
\texttt{oftg\_curve\_grpo\_s3} & B & 1 & 11.8 \\
\texttt{oftg\_curve\_bc\_s1} & A & 1 & 15.5 \\
\texttt{oftg\_curve\_bct\_s1} & A & 1 & -- \\
\texttt{oftg\_curve\_grpo\_s1} & A & 1 & 14.8 \\
\midrule
Total on record & & & 213.4 \\
\bottomrule
\end{tabular}}
\end{table}

\section{Example rollouts}
\label{sec:rollouts}
\begin{figure*}[t]
  \centering
  \includegraphics[width=0.86\textwidth]{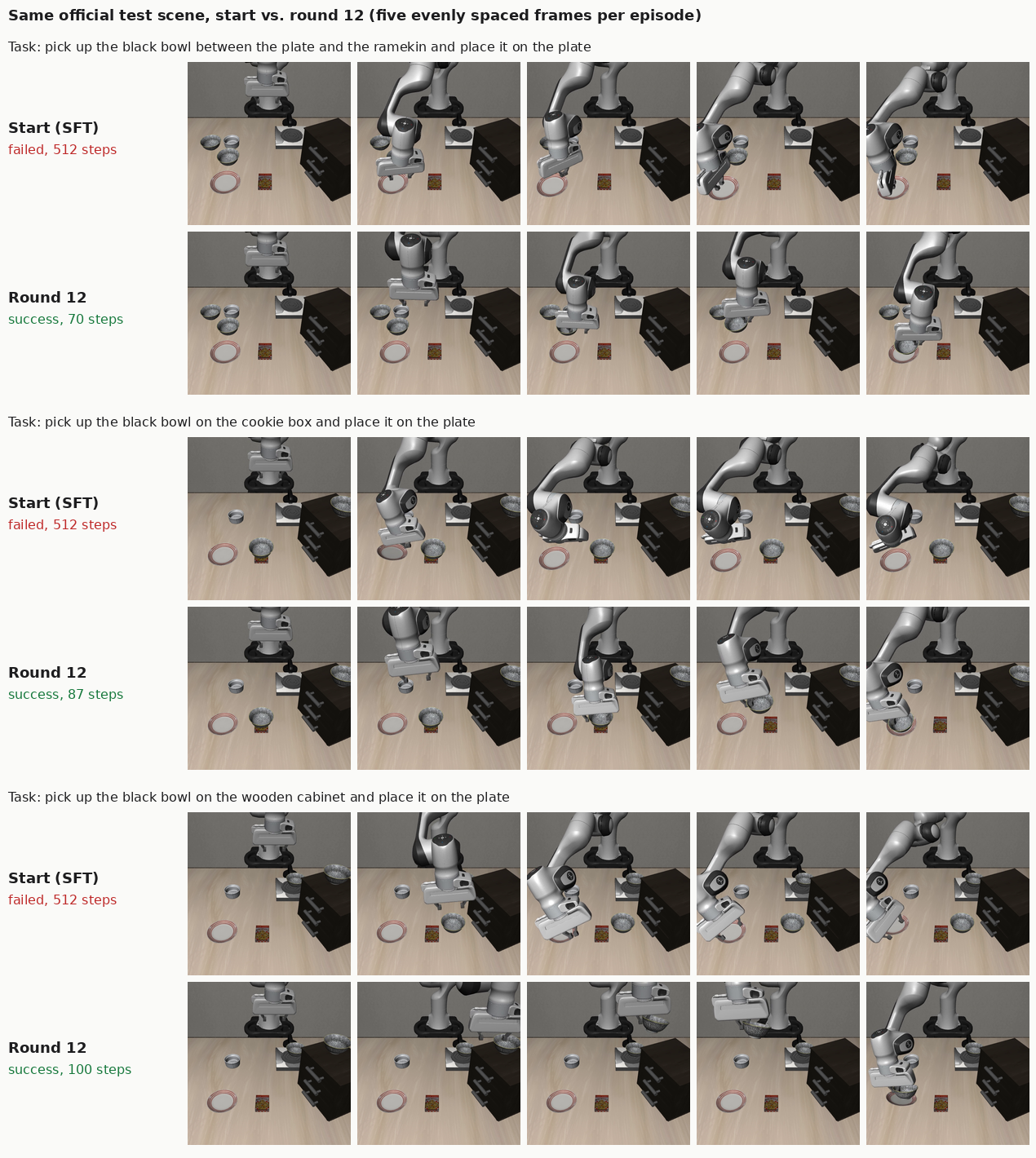}
  \caption{\textbf{Example rollouts on official LIBERO-Spatial test scenes.} For each scene, the starting policy (top) and the policy retained after twelve rounds (bottom), five evenly spaced frames per episode; the label gives the simulator's result and episode length. Hand-picked illustration (\cref{sec:rollouts}).}
  \label{fig:rollouts}
\end{figure*}

The rollout video on the project page shows five official LIBERO-Spatial test scenes, five different tasks, with the starting policy and with the policy retained after twelve rounds of the recursive loop with the fixed GRPO program (seed 0, platform A; retained at round 12 with gate success 75.5\%); \cref{fig:rollouts} shows three of them. The scenes were chosen by hand among the 41 official states where the recorded test outcome changed from failure (start) to success (round 12), preferring short successes and distinct tasks, and were rendered again for the video with the official test protocol (greedy actions); six of eight re-rendered states reproduced the recorded success, and the five with distinct tasks are shown. They illustrate behaviour and are not a measure of success rates, which are in the paper's tables.

\section{Code, preregistration and records}
\label{sec:code}
The code release contains the framework (engine, gate, LIBERO domain, OpenVLA-OFT rollout and update workers, numeric and LLM proposers), the registered analysis scripts that produce every number in the paper and in this supplement, the preregistration and its dated addenda, and the records of every run in the reported studies (round records, retained programs, test outcomes per episode, LLM proposer logs). Platform-specific launch scripts are omitted, and names of people, machines and storage locations are removed. Its \texttt{README.md} maps each table and figure to the script and record that produce it; \texttt{scripts/supplement/} regenerates every table of this supplement from the records, including the Goal fresh-scene meta-level effect of \cref{tab:components}.

\end{document}